\documentclass[11pt]{article}

\usepackage[final]{acl}

\usepackage{times}
\usepackage{latexsym}

\usepackage[T1]{fontenc}

\usepackage[utf8]{inputenc}

\usepackage{microtype}

\usepackage{inconsolata}

\usepackage{graphicx}

\usepackage{algorithm}
\usepackage{algorithmic}
\usepackage{xcolor}
\usepackage{booktabs}
\usepackage{amsmath}

\usepackage{colortbl}
\usepackage{multirow}
\usepackage{enumitem}

\title{Authorship Attribution in Multilingual Machine-Generated Texts}

\author{Lucio {La Cava}\textsuperscript{\rm 1},
     Dominik Macko\textsuperscript{\rm 2},
     Robert Moro\textsuperscript{\rm 2},
     Ivan Srba\textsuperscript{\rm 2},
     Andrea Tagarelli\textsuperscript{\rm 1} \\
     \textsuperscript{\rm 1}DIMES Department, University of Calabria, Italy\\
     \textsuperscript{\rm 2}Kempelen Institute of Intelligent Technologies, Slovakia\\
     \texttt{\{lucio.lacava, tagarelli\}@dimes.unical.it}\\
     \texttt{\{dominik.macko, robert.moro, ivan.srba\}@kinit.sk}\\}

\newtheorem{problem}{Problem}
\def\X{\mathcal{X}}
\def\Y{\mathcal{Y}}
\def\L{\mathcal{L}}
\def\M{\mathcal{M}}
\def\MMGT{\textsf{ML-MGT}}
\def\CLMGT{\textsf{CL-MGT}}
\def\multitude{\textsc{MULTITuDE}}

\begin{document}

\maketitle

\begin{abstract}
As Large Language Models (LLMs) have reached human-like fluency and coherence, distinguishing machine-generated text (MGT) from human-written content becomes increasingly difficult. While early efforts in MGT detection have focused on binary classification, the growing landscape and diversity of LLMs require a more fine-grained yet challenging \textit{authorship attribution} (AA), i.e., being able to identify the precise generator (LLM or human) behind a text.  
However, AA remains nowadays confined to a monolingual setting, with English being the most investigated one, overlooking the multilingual nature and usage of modern LLMs. 
In this work, we introduce the problem of \textit{Multilingual Authorship Attribution}, which involves attributing texts to human or multiple LLM generators across diverse languages. 
Focusing on 18 languages---covering multiple families and writing scripts---and 8 generators (7 LLMs and the human-authored class), we investigate the multilingual suitability of monolingual AA methods   in terms of  their cross-lingual transferability, and the impact of   generators on attribution performance. 
Our results reveal that while certain monolingual AA methods can be adapted to multilingual settings, significant limitations and challenges remain, particularly in transferring across diverse language families,   
underscoring the complexity of multilingual AA and the need for more robust approaches to better match real-world scenarios. 
\end{abstract}

\section{Introduction}
\label{intro}

Large Language Models (LLMs) have nowadays reached a level of fluency and coherence that enables them to produce human-like text that is no longer distinguishable from that written by humans~\cite{Jakesch2023human}. 
While these advancements pave the way for new opportunities in communication, creativity, and productivity~\cite{bubeck2023sparks}, they also raise critical risks in our society about transparency, accountability, and misuse. In particular, the inability to effectively determine whether a text has been generated by humans or machines leaves many open risks for our society, such as misinformation~\cite{Chen24iclr}, disinformation~\cite{zugecova-etal-2025-evaluation}, and copyright infringement~\cite{liu-etal-2024-shield}. 

Early attempts to address the above  challenge  focused on binary \textit{machine-generated text} (MGT) detection, i.e., automated approaches for distinguishing  (AI-based) synthetically generated  text from human-written text. However, while effective in many contexts, binary detection has faced a strong limitation: the inability to account for a growing diversity of LLM generators. Indeed, as the number of released LLMs continues to expand day by day, so does the need for fine-grained \textit{authorship attribution} (AA): not just identifying that a text is machine-generated, but also determining which model produced it~\cite{uchendu-etal-2020-authorship}.% 

Despite this, existing attempts to perform authorship attribution remain confined to a monolingual setting---with English being the most prominent. This represents a critical blind spot, since modern LLMs are increasingly multilingual, trained to generate content in a broad range of languages, and used in diverse linguistic and cultural contexts. 

To address this gap, in this work, we define and investigate the problem of \textit{multilingual authorship attribution}, i.e., attributing texts to the corresponding generators (being they LLMs or humans), across multiple languages and writing scripts. In particular, our study aims to evaluate the multilingual suitability and cross-lingual generalizability of existing AA approaches in this challenging setting, through the following research questions:\\
\textbf{RQ1} --- \textit{How effectively can existing authorship attribution methods handle multilingual machine-generated text (\MMGT{})?}\\
\textbf{RQ2} --- \textit{To what extent can authorship attribution approaches for \MMGT{} transfer across different languages and language families?}\\
\textbf{RQ3} --- \textit{How does the choice of generator model influence the multilingual suitability and cross-lingual generalizability of authorship attribution methods?}

\vspace{1.5mm}
\noindent\textbf{Contributions.\ }
By answering these research questions, our contributions in this work are as follows:

\begin{itemize} \itemsep=1pt
    \item We introduce and formally define the problems of \MMGT{} and Cross-lingual Machine-generated Text (\CLMGT{})  Authorship Attribution, which handle  attributing texts to their machine/human generators across multiple languages and families.
    \item We evaluate the {suitability of existing monolingual authorship attribution methods} to the multilingual setting, analyzing how well current monolingual approaches perform in this more challenging scenario, covering {18 languages} and {8 different generators}.
    \item We investigate the {cross-lingual transferability}
    of authorship attribution methods, assessing their robustness when applied to previously unseen languages.
\end{itemize}

Our findings suggest that while most existing authorship attribution methods can be extended to the multilingual setting, with varying degrees of efficacy, several challenges persist. Indeed, current authorship attribution methods struggle to generalize across dissimilar language families or writing scripts, as performances are  heavily affected by the linguistic properties of the target languages and the identity of the generators. 
These points underscore the challenges introduced by our newly defined \MMGT{} and \CLMGT{} problems and highlight the pressing need to develop more robust, language-agnostic attribution methods capable of handling the linguistic and stylistic diversity present in real-world multilingual scenarios.

\section{Related Work}
\label{sec:related}
The human-like text generation capabilities achieved by LLMs in recent years have blurred the distinction between human-authored and machine-generated texts, intensifying the need for reliable detection methods~\cite{jawahar-etal-2020-automatic, crothers2023machine, tang2024science, wu-etal-2025-survey}.

\vspace{1.5mm}
\noindent\textbf{MGT Detection.\ } 
In response to this challenge, we witnessed a surge in the development of detection methods. These include statistical learning approaches such as probabilistic modeling~\cite{mitchell2023detectgpt, bao2023fast, wang-etal-2023-seqxgpt,miao-etal-2024-efficient}, log-rank~\cite{su-etal-2023-detectllm} and perplexity-based methods~\cite{vasilatos2023howkgpt}, and stylistic or discourse-based approaches~\cite{kim-etal-2024-threads,gehrmann2019gltr, tulchinskii2023intrinsic, venkatraman-etal-2024-gpt}. Also, watermarking techniques were developed to embed signals in generated texts that remain invisible to humans but are algorithmically detectable~\cite{pmlr-v202-kirchenbauer23a, yoo-etal-2023-robust, xu2024learning} for post-hoc detection. More recently, learning-based methods have gained traction, including deep neural classifiers~\cite{ippolito-etal-2020-automatic, verma-etal-2024-ghostbuster}, contrastive learning frameworks~\cite{bhattacharjee-etal-2023-conda,bhattacharjee2024eagle}, the use of ChatGPT itself as a detector~\cite{Bhattacharjee2024gpt}, and hybrid approaches incorporating topological features~\cite{uchendu2023toproberta}.

\vspace{1.5mm}
\noindent\textbf{MGT Authorship Attribution.\ }
As the diversity of generative models continues to grow, researchers have begun shifting their focus from mere detection to the more ambitious task of \textit{authorship attribution}. This task requires identifying which specific model produced a given text~\cite{uchendu-etal-2020-authorship}, with important implications for accountability, provenance tracking, and mitigation of misuse~\cite{huang2025surveyattribution, Uchendu2023aa}.

Early works explored the possibility of attributing texts to generators through statistical signals~\cite{solaiman2019release, gehrmann2019gltr}, but fell short in performance as shown in~\cite{la-cava-tagarelli-2025-openturingbench}. More recent approaches adopt deep learning and contrastive learning strategies, showing stronger results in controlled settings~\cite{guo2024detective, LaCava2024ecai, He2024mgtbench}. Nevertheless, the body of work on attribution is relatively limited compared to detection.

\vspace{1.5mm}
\noindent\textbf{Multilingual MGT Authorship Attribution.\ }
Despite growing attention to attribution, the entire line of research remains fundamentally monolingual, with a predominant focus on English~\cite{wang-etal-2024-semeval-2024, LaCava2024ecai}. A handful of studies have extended to Russian~\cite{Shamardina_2022} and Spanish~\cite{sarvazyan2023overview}, but a systematic investigation of multilingual attribution and the related impact of languages remains underexplored.

This lack   motivates our work, and the investigation of multilingual authorship attribution and cross-lingual transferability of attribution methods, as formalized next.

\section{Problem Statement}
\label{sec:problem-statement}

Let us denote with $\L$ a set of \textit{languages} and with   $\M$ a set of \textit{machine generators}, i.e., LLMs producing MGTs. 
 \textit{Authorship attribution of multilingual machine-generated text} (\MMGT) can be formulated as a multi-class classification problem,  defined as follows. 

\begin{problem}[\MMGT]\label{p1}
We are given a set of texts $\X = \X_h \cup \X_m$, consisting of two subsets: $\X_h$, which contains human-written texts, and $\X_m$, which contains machine-generated texts (MGTs) from all models in $\M$. Each text in $\X_h$ and $\X_m$ is written in a language from the set $\L$. Accordingly, we  express these subsets as $\X_h = \bigcup_{\ell \in \L} \X_{h,\ell}$ and $\X_m = \bigcup_{\ell \in \L} \X_{m,\ell}$, where $\X_{h,\ell}$ and $\X_{m,\ell}$ denote the human-written and MGTs in language $\ell$, respectively.
 
If we denote with $y_h$ the `\textsc{Human}' class label and with $\Y_m = \{y_j\}_{j=1}^{|\M|}$  the set of `\textsc{Machine}' class labels, the task  is to   recognize
the author of a given text choosing among the human ($y_{h}$) and the machine generators in $\M$,  i.e., to learn a mapping function $f:\widehat{\X} \mapsto \Y=\{y_{h}\}  \cup \Y_m$, with   $\widehat{\X} \subseteq \X$.     
\end{problem}

In Problem \ref{p1}, the choice of $\widehat{\X} = \widehat{\X}_h \cup \widehat{\X}_m$ relies on the definition of a \textit{language-selection strategy} $g(\cdot)$ such that, for any $L', L'' \subseteq \L$, 
$\widehat{\X}_h = g(\X_h, L')$ and 
$\widehat{\X}_m = g(\X_m, L'')$ are the subsets of $\X_h$, resp. $\X_m$, which select the texts written in any language in  $L'$, resp. $L''$. 
Unless otherwise specified, we hereinafter assume that $L'=L''$, which implies   that \textit{human-written texts and MGTs are provided in the same languages and aligned in a pairwise fashion.}

\begin{problem}[\CLMGT]
Let $\L_{train} \subseteq \L$ be the set of languages used for training $f$, and   $\L_{test} \subseteq \L$ be the set of test languages. 
Problem \ref{p1} reduces to an instance of \emph{cross-lingual transferability}    if $\L_{train} \subset \L_{test}$.
\end{problem}

The   cross-lingual transferability problem aims  to evaluate how well a model trained on a set of   \textit{source} languages can generalize to \textit{target} languages that were not seen during training. If the test set includes additional languages not seen during training, then the model must rely on its ability to transfer knowledge across languages.

\begin{table}[t]
\centering
\setlength{\tabcolsep}{3pt}
\scalebox{0.8}{
\begin{tabular}{llcrr}
\toprule
\textbf{Family} & \textbf{Language} & \textbf{Code} & \textbf{Train} & \textbf{Test} \\
\midrule 
\multirow{3}{*}{Germanic} & 
Dutch        & nl & 7958 & 2386 \\
& English      & en & 7954 & 2384 \\
& German       & de & 7951 & 2388 \\
\midrule
Hellenic & 
Greek        & el & 7944 & 2384 \\
\midrule 
Semitic & 
Arabic       & ar & 7975 & 2392 \\
\midrule 
Sino-Tibetan & 
Chinese      & zh & 7926 & 2383 \\
\midrule 
\multirow{3}{*}{Slavic-Cyrillic} & 
Bulgarian    & bg & 7954 & 2386 \\
& Ukrainian    & uk & 7939 & 2385 \\
& Russian      & ru & 7945 & 2382 \\
\midrule 
\multirow{5}{*}{Slavic-Latin}    & Croatian     & hr & 7951 & 2384 \\
& Czech        & cs & 7962 & 2389 \\
 & Polish       & pl & 7946 & 2383 \\
  & Slovak       & sk & 7946 & 2385 \\
    & Slovenian    & sl & 7947 & 2386 \\
\midrule  
\multirow{3}{*}{Romanic} & 
Portuguese   & pt & 7956 & 2388 \\
& Romanian     & ro & 7949 & 2386 \\
& Spanish      & es & 7947 & 2387 \\
 \midrule 
Uralic & 
Hungarian    & hu & 7964 & 2385 \\
\midrule
\textbf{Total} & -- & -- & \textbf{143,114} & \textbf{42,943} \\
\bottomrule
\end{tabular}
}
\caption{Per-language sample counts for train/test splits of the selected data from the \multitude{} dataset.}
\label{tab:multitude-stats}
\vspace{-3mm}
\end{table}

\section{Data and Generator Models}
\label{sec:data}

To conduct our study, we resorted to the \multitude{} (v3) dataset~\cite{macko_2025_15519413,macko-etal-2025-multisocial}. It contains LLM-generated and human-written news articles, where the latter come from the \textit{MassiveSum} collection~\cite{varab-schluter-2021-massivesumm}.
The machine-generated counterparts are generated by seven LLMs prompted with the original headlines of the articles. These LLMs cover a representative body of open and commercially licensed families of models, spanning various model sizes, architectures, and pre-training strategies, namely \textit{Mistral-7B-Instruct-v0.2}, \textit{OPT-IML-Max-30B}, \textit{v5-Eagle-7B-HF}, \textit{Vicuna-13B}, \textit{Llama-2-70B-Chat-HF}, \textit{Aya-101}, and \textit{GPT-3.5-Turbo-0125}.

Our choice over other existing multilingual MGT datasets (cf. Appendix~\ref{app:ml-datasets}), such as M4GT-Bench~\cite{wang-etal-2024-m4gt} or RAID~\cite{dugan-etal-2024-raid}, was driven by the consistent set of generators, text-generation settings, and domains for each language, enabling focus on unbiased cross-lingual transferability aspects.

Among the 21 languages available in \multitude, we focused on the 18 languages that (i) provide fully balanced coverage across language-generator combinations, to avoid skewed or underrepresented distributions that could bias evaluation, and (ii) contain at least 95\% of the target number of samples---1,000 per generator for training and 300 per generator for testing---for a  robust and fair comparison across languages and models. 
Table~\ref{tab:multitude-stats} provides   statistics     on the train-test splits. Each value reflects a uniform distribution across all classes, with 1/8 of the samples assigned to human-written texts, and the remainder evenly distributed across the    LLM generators.

\begin{figure}[t!]
\centering
\includegraphics[width=\linewidth]{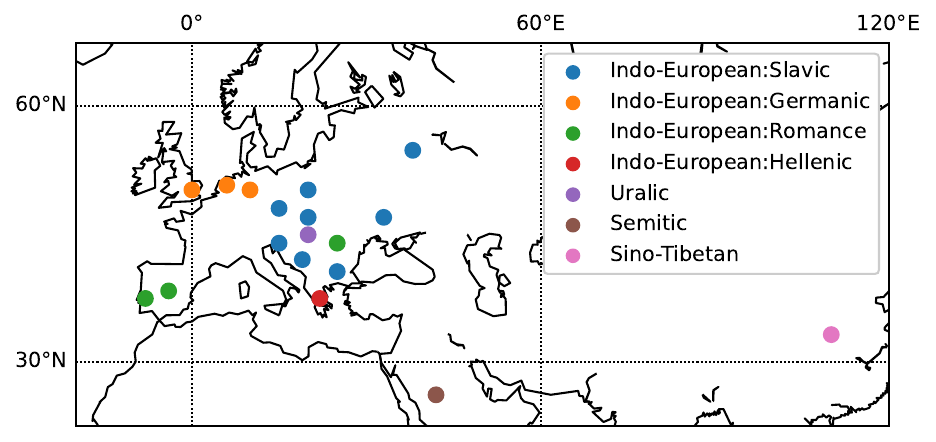}
\caption{Language coverage for our multilingual AA.}
\label{fig:multitude-map}
\end{figure}

\paragraph{Language analysis.}  
As shown in Fig.~\ref{fig:multitude-map}, our selected data covers eight language families, namely 
Indo-European---organized into 
Germanic, Romanic, Slavic-Latin,  Slavic-Cyrillic, and Hellenic---Uralic, Semitic, and Sino-Tibetan. This also corresponds to  five writing scripts ($12\times$Latin, $3\times$Cyrillic, $1\times$Arabic, $1\times$Hanzi, and $1\times$Greek). % 
Thus, the selected language composition enables various combinations of investigations and in-depth insights regarding multilingual and cross-lingual characteristics of AA methods.

\section{Detection Methods}
\label{sec:methods}

In this section,  we present the methods selected for evaluation in a multilingual setting.  These were chosen based on their strong performance in recent works~\cite{sarvazyan2023overview, He2024mgtbench, wang-etal-2024-semeval-2024, la-cava-tagarelli-2025-openturingbench}. 
However, all of them required adaptation to suit the specific demands of our target attribution problems, i.e., \MMGT{} and \CLMGT{}. Next, we detail the adaptation process for each method.

\subsection{Statistical Approaches}
\noindent\textbf{Individual statistical approaches.\ } 
We consider two zero-shot binary detectors to extract statistical features from texts, i.e., Fast-DetectGPT~\cite{bao2023fast} with mGPT-13B\footnote{mGPT is a multilingual model with a similar architecture to the default GPT-J/Neo, outperforming other tested models in our experiments, including XLM-R, Qwen3-4B and 14B.} by~\citealp{shliazhko-etal-2024-mgpt} as both
the reference and sampling model, and Binoculars~\cite{hans2024spottingllmsbinocularszeroshot} with Falcon-7B by~\citealp{falcon40b} as an observer model and Falcon-7B-Instruct as a
performer model. Following~\cite{He2024mgtbench, spiegel-macko-2024-imgtb, la-cava-tagarelli-2025-openturingbench}, we train a Logistic Regressor on top of the extracted features to perform multiclass classification for the AA task.

\vspace{1.5mm}
\noindent\textbf{Ensemble statistical approaches.\ }
To provide a stronger statistical approach, we   combine nine statistical features into a statistical ensemble dubbed \textsl{StatEnsemble}. These are the metrics of Binoculars~\cite{hans2024spottingllmsbinocularszeroshot}, Fast-DetectGPT~\cite{bao2023fast}, perplexity, Rank~\cite{gehrmann2019gltr}, log-rank, log-likelihood, Entropy~\cite{10.5555/3053718.3053722}, LLM-Deviation~\cite{wu2023mfd}, and DetectLLM-LRR~\cite{su-etal-2023-detectllm}, calculated based on mGPT-13B outputs. To perform a multiclass classification for AA, we train a Multi-layer Perceptron (MLP) classifier (MLP performing the best out of the examined Logistic Regressor, MLP, and Random Forest) with hyperparameters optimized using 5-fold grid search cross-validation over 1,000 steps. The remainder is kept to the default values of the \texttt{scikit-learn} library we used.

\subsection{LLM-based Supervised Approaches}
\noindent\textbf{Fine-tuned encoders.\ }
For this type of detector, we consider RoBERTa-large~\cite{liu2019robertarobustlyoptimizedbert} as an English-only pre-trained language model, and XLM-RoBERTa-large~\cite{DBLP:journals/corr/abs-1911-02116} as the multilingual counterpart. Both models were fine-tuned for the AA task following~\cite{wang-etal-2024-semeval-2024,sarvazyan2023overview}, with a learning rate of $2e\text{-}6$ and max sequence length of $512$ tokens.

\vspace{1.5mm}
\noindent\textbf{Contrastive learner.\ }
As a representative of contrastive approaches, we adapt the OTBDetector~\cite{la-cava-tagarelli-2025-openturingbench}, which serves as the best-performing method in the recent \textit{OpenTuringBench} benchmark for MGT attribution, to the multilingual AA task. It uses contrastive learning for fine-tuning a pre-trained model to separate latent representations of texts from different generators. For the multilingual setting, we replaced the original Longformer model with XLM-RoBERTa-large to ensure multilingual generalizability.

\vspace{1.5mm}
\noindent\textbf{Fine-tuned decoder.\ }
We adapt the   \textsl{mdok} detector~\cite{macko2025mdokkinitrobustlyfinetuned},     originally conceived as a multilingual binary MGT detection method, to the multilingual AA task. It is based on a fine-tuning of Qwen3-4B-Base~\cite{qwen3technicalreport} model via QLoRA for enhancing generalization to out-of-distribution and obfuscated data, with a multiclass classification head performing multilingual classification. %  
In addition, we have included Qwen3-4B-Base itself, fine-tuned similarly as the encoder models above.

\section{Experimental Setup}
\label{sec:experimental}

To address our research questions, we design four tasks that evaluate the feasibility and generalizability of multilingual authorship attribution. %  
The first task  corresponds to solving the \MMGT{} problem (\textbf{RQ1}). % 
To address the \CLMGT{} problem (\textbf{RQ2}), we distinguish between \textit{per-language} and  \textit{per-language-family} cross-lingual transferability. 
The latter task corresponds to   investigate   the impact of the various LLM generators   on the \MMGT{} and \CLMGT{} performance (\textbf{RQ3}).

\vspace{2mm}
\noindent\textbf{RQ1. Suitability of Existing Approaches to \MMGT{}.\ }
To address {RQ1}, we evaluate  the ability of the selected methods to handle the \MMGT{} problem, by training them  on data from all   languages jointly, covering all 8 classes (7 LLM generators and human-authorship). The  multilingual test set comprises the same languages, with performance reported as the macro-averaged $F_1$ score across all classes to ensure balanced treatment of each class regardless of frequency. 
Details on the train/test splits  are shown in Table~\ref{tab:multitude-stats}.

\vspace{2mm}
\noindent\textbf{RQ2. Cross-lingual transferability of \MMGT{} Authorship Attribution.\ }
We investigate  whether AA methods trained on a single language or a combination of multiple languages could generalize their capabilities to other   languages.  

First, following~\cite{macko-etal-2023-multitude,macko-etal-2024-authorship,macko-etal-2025-multisocial}, 
we train AA methods on the subsets of English-, Spanish-, and Russian-only data from Table~\ref{tab:multitude-stats}, using all 8 classes. Additionally, we  train AA methods on a combination of English, Spanish, and Russian train data, which are sampled to $1/3$ each to ensure that these methods are trained on the same number of training samples as the monolingually trained AA methods.  % 
It should be noted that our choice of English, Spanish and Russian  is motivated since they are the most popular languages with the two most representative scripts in \multitude{}, i.e., Latin and Cyrillic.

To assess the \textit{per-language transferability}, we evaluate the macro-averaged $F_1$ of AA methods on all the languages (including English, Spanish, and Russian), thus  examining how  a single language or   language-subset during training can steer detectors to perform well in other  languages,   and comparing them  to the multilingually trained detectors.  

Similarly, to assess the \textit{per-language-family transferability}, we  investigate how the methods trained on one writing script  can generalize to languages using a different script, hence to understand whether a language family plays a role in cross-lingual generalization. 
To this purpose, again we  use English and Spanish to represent Latin-script training and Russian to represent Cyrillic-script training, and perform evaluation  across all   languages, measuring macro-averaged~$F_1$.

By comparing intra-family and inter-family 
transfer performance, we aim to quantify whether 
family similarity/divergence affects the transferability of current AA   methods in multilingual settings.

\vspace{2mm}
\noindent\textbf{RQ3. Impact of LLM generators on the    \MMGT{} and   \CLMGT  performance.\ }
We explore how the LLM generators influence the \MMGT{} performance and   cross-lingual generalization of attribution methods. % 
 To this aim, we examine language variations in class-level $F_1$ scores for each generator, shedding light on the interplay between generator identity and linguistic context in shaping adaptability and transferability.

\begin{table*}[!t]
\centering
\setlength{\tabcolsep}{1.0mm} 
\scalebox{0.78}{
\begin{tabular}{l||c|c|c||c|c|c||c|c|c|c|c||c|c|c||c|c|c|c||c}
Lang. family $\rightarrow$ & \multicolumn{3}{c||}{Germanic} & \multicolumn{3}{c||}{Romance} & \multicolumn{5}{c||}{Slavic-Latin} & \multicolumn{3}{c||}{Slavic-Cyrillic} & \multicolumn{4}{c||}{Others} & \\
Method $\downarrow$ & de & en & nl & es & pt & ro & cs & hr & pl & sk & sl & bg & ru & uk & hu & el & ar & zh & all \\
\midrule
Qwen3-4B-Base & \bfseries {\cellcolor[HTML]{98D486}}0.92 & {\cellcolor[HTML]{9AD587}}0.91 & \bfseries {\cellcolor[HTML]{90D083}}0.95 & \bfseries {\cellcolor[HTML]{98D486}}0.92 & \bfseries {\cellcolor[HTML]{95D385}}0.93 & \bfseries {\cellcolor[HTML]{92D183}}0.95 & \bfseries {\cellcolor[HTML]{90D083}}0.96 & {\cellcolor[HTML]{92D183}}0.95 & {\cellcolor[HTML]{93D284}}0.94 & {\cellcolor[HTML]{8ED082}}0.97 & {\cellcolor[HTML]{90D083}}0.95 & \bfseries {\cellcolor[HTML]{93D284}}0.95 & \bfseries {\cellcolor[HTML]{98D486}}0.92 & {\cellcolor[HTML]{97D385}}0.93 & \bfseries {\cellcolor[HTML]{95D385}}0.93 & {\cellcolor[HTML]{95D385}}0.93 & \bfseries {\cellcolor[HTML]{8ED082}}0.96 & {\cellcolor[HTML]{A9DB8C}}0.85 & \bfseries {\cellcolor[HTML]{95D385}}0.93 \\
mdok & {\cellcolor[HTML]{9AD587}}0.92 & \bfseries {\cellcolor[HTML]{9AD587}}0.91 & {\cellcolor[HTML]{92D183}}0.95 & {\cellcolor[HTML]{9CD687}}0.91 & {\cellcolor[HTML]{97D385}}0.93 & {\cellcolor[HTML]{93D284}}0.94 & {\cellcolor[HTML]{92D183}}0.95 & \bfseries {\cellcolor[HTML]{90D083}}0.96 & \bfseries {\cellcolor[HTML]{93D284}}0.94 & \bfseries {\cellcolor[HTML]{8DCF81}}0.97 & \bfseries {\cellcolor[HTML]{90D083}}0.95 & {\cellcolor[HTML]{95D385}}0.93 & {\cellcolor[HTML]{9AD587}}0.91 & \bfseries {\cellcolor[HTML]{95D385}}0.93 & {\cellcolor[HTML]{97D385}}0.93 & \bfseries {\cellcolor[HTML]{93D284}}0.94 & {\cellcolor[HTML]{90D083}}0.96 & \bfseries {\cellcolor[HTML]{A4D98A}}0.87 & {\cellcolor[HTML]{95D385}}0.93 \\
OTBDetector & {\cellcolor[HTML]{A2D88A}}0.87 & {\cellcolor[HTML]{B8E293}}0.78 & {\cellcolor[HTML]{9AD587}}0.91 & {\cellcolor[HTML]{A7DB8C}}0.85 & {\cellcolor[HTML]{A1D889}}0.89 & {\cellcolor[HTML]{95D385}}0.93 & {\cellcolor[HTML]{95D385}}0.93 & {\cellcolor[HTML]{95D385}}0.93 & {\cellcolor[HTML]{98D486}}0.92 & {\cellcolor[HTML]{90D083}}0.96 & {\cellcolor[HTML]{93D284}}0.94 & {\cellcolor[HTML]{97D385}}0.93 & {\cellcolor[HTML]{A4D98A}}0.87 & {\cellcolor[HTML]{9CD687}}0.91 & {\cellcolor[HTML]{9CD687}}0.91 & {\cellcolor[HTML]{98D486}}0.92 & {\cellcolor[HTML]{93D284}}0.95 & {\cellcolor[HTML]{B3E091}}0.80 & {\cellcolor[HTML]{9DD688}}0.90 \\
XLM-R-large & {\cellcolor[HTML]{B2DF90}}0.81 & {\cellcolor[HTML]{CFEC9E}}0.65 & {\cellcolor[HTML]{ABDC8D}}0.84 & {\cellcolor[HTML]{BAE394}}0.76 & {\cellcolor[HTML]{B2DF90}}0.80 & {\cellcolor[HTML]{A4D98A}}0.87 & {\cellcolor[HTML]{A1D889}}0.88 & {\cellcolor[HTML]{A1D889}}0.88 & {\cellcolor[HTML]{A2D88A}}0.88 & {\cellcolor[HTML]{97D385}}0.93 & {\cellcolor[HTML]{9DD688}}0.90 & {\cellcolor[HTML]{A4D98A}}0.87 & {\cellcolor[HTML]{B6E192}}0.78 & {\cellcolor[HTML]{ABDC8D}}0.84 & {\cellcolor[HTML]{A6DA8B}}0.86 & {\cellcolor[HTML]{A2D88A}}0.88 & {\cellcolor[HTML]{9DD688}}0.90 & {\cellcolor[HTML]{C1E698}}0.72 & {\cellcolor[HTML]{ABDC8D}}0.84 \\
RoBERTa-large & {\cellcolor[HTML]{B6E192}}0.78 & {\cellcolor[HTML]{C1E698}}0.72 & {\cellcolor[HTML]{B1DF90}}0.81 & {\cellcolor[HTML]{BDE496}}0.74 & {\cellcolor[HTML]{B2DF90}}0.80 & {\cellcolor[HTML]{A9DB8C}}0.84 & {\cellcolor[HTML]{AEDD8E}}0.83 & {\cellcolor[HTML]{ACDD8E}}0.83 & {\cellcolor[HTML]{B1DF90}}0.81 & {\cellcolor[HTML]{A9DB8C}}0.85 & {\cellcolor[HTML]{ABDC8D}}0.84 & {\cellcolor[HTML]{D2EDA0}}0.63 & {\cellcolor[HTML]{D3EDA0}}0.63 & {\cellcolor[HTML]{CCEA9D}}0.67 & {\cellcolor[HTML]{BAE394}}0.76 & {\cellcolor[HTML]{DAF0A4}}0.59 & {\cellcolor[HTML]{C5E89A}}0.70 & {\cellcolor[HTML]{D9F0A3}}0.60 & {\cellcolor[HTML]{BDE496}}0.75 \\
StatEnsemble & {\cellcolor[HTML]{E7F6AD}}0.49 & {\cellcolor[HTML]{F8FCC0}}0.33 & {\cellcolor[HTML]{DEF2A7}}0.55 & {\cellcolor[HTML]{ECF7B1}}0.45 & {\cellcolor[HTML]{E9F6AF}}0.47 & {\cellcolor[HTML]{E8F6AE}}0.48 & {\cellcolor[HTML]{EDF8B2}}0.43 & {\cellcolor[HTML]{EEF9B3}}0.43 & {\cellcolor[HTML]{E5F5AC}}0.50 & {\cellcolor[HTML]{EEF9B3}}0.43 & {\cellcolor[HTML]{F9FDC2}}0.31 & {\cellcolor[HTML]{E3F4AA}}0.51 & {\cellcolor[HTML]{E8F6AE}}0.48 & {\cellcolor[HTML]{E8F6AE}}0.48 & {\cellcolor[HTML]{E5F5AC}}0.50 & {\cellcolor[HTML]{F0F9B4}}0.41 & {\cellcolor[HTML]{F2FAB5}}0.40 & {\cellcolor[HTML]{F7FCBC}}0.35 & {\cellcolor[HTML]{ECF7B1}}0.45 \\
Fast-DetectGPT & {\cellcolor[HTML]{FBFDCE}}0.25 & {\cellcolor[HTML]{FFFFE5}}{\cellcolor{white}}0.12 & {\cellcolor[HTML]{FBFDCF}}0.25 & {\cellcolor[HTML]{FDFEDB}}0.18 & {\cellcolor[HTML]{FDFED9}}0.20 & {\cellcolor[HTML]{FDFEDA}}0.19 & {\cellcolor[HTML]{FCFED3}}0.23 & {\cellcolor[HTML]{FCFED4}}0.22 & {\cellcolor[HTML]{FAFDCB}}0.26 & {\cellcolor[HTML]{FDFEDB}}0.18 & {\cellcolor[HTML]{FCFED7}}0.20 & {\cellcolor[HTML]{F9FDC2}}0.31 & {\cellcolor[HTML]{F9FDC2}}0.31 & {\cellcolor[HTML]{F9FDC2}}0.31 & {\cellcolor[HTML]{F9FDC5}}0.30 & {\cellcolor[HTML]{FEFFDF}}0.16 & {\cellcolor[HTML]{FDFEDD}}0.17 & {\cellcolor[HTML]{FEFFDF}}0.16 & {\cellcolor[HTML]{FBFED2}}0.23 \\
Binoculars & {\cellcolor[HTML]{FCFED7}}0.20 & {\cellcolor[HTML]{FEFFE1}}0.15 & {\cellcolor[HTML]{FCFED4}}0.22 & {\cellcolor[HTML]{FEFFE1}}0.15 & {\cellcolor[HTML]{FDFEDD}}0.18 & {\cellcolor[HTML]{FBFED0}}0.24 & {\cellcolor[HTML]{FEFFE2}}0.14 & {\cellcolor[HTML]{FEFFE2}}0.14 & {\cellcolor[HTML]{FCFED3}}0.23 & {\cellcolor[HTML]{FFFFE5}}0.13 & {\cellcolor[HTML]{FEFFDE}}0.17 & {\cellcolor[HTML]{FFFFE5}}{\cellcolor{white}}0.07 & {\cellcolor[HTML]{FFFFE5}}0.13 & {\cellcolor[HTML]{FFFFE5}}{\cellcolor{white}}0.08 & {\cellcolor[HTML]{FFFFE5}}0.13 & {\cellcolor[HTML]{FEFFE2}}0.14 & {\cellcolor[HTML]{FFFFE5}}{\cellcolor{white}}0.12 & {\cellcolor[HTML]{FFFFE4}}0.14 & {\cellcolor[HTML]{FEFFDF}}0.16 \\
\midrule
\textit{Average} & {\cellcolor[HTML]{CEEB9E}}0.65 & {\cellcolor[HTML]{DCF1A5}}0.57 & {\cellcolor[HTML]{C8E99B}}0.68 & {\cellcolor[HTML]{D5EEA1}}0.62 & {\cellcolor[HTML]{CFEC9E}}0.65 & {\cellcolor[HTML]{C9E99C}}0.68 & {\cellcolor[HTML]{CBEA9C}}0.67 & {\cellcolor[HTML]{CBEA9C}}0.67 & {\cellcolor[HTML]{C8E99B}}0.69 & {\cellcolor[HTML]{C9E99C}}0.68 & {\cellcolor[HTML]{CEEB9E}}0.66 & {\cellcolor[HTML]{CFEC9E}}0.65 & {\cellcolor[HTML]{D3EDA0}}0.63 & {\cellcolor[HTML]{D0EC9F}}0.64 & {\cellcolor[HTML]{CCEA9D}}0.66 & {\cellcolor[HTML]{D5EEA1}}0.62 & {\cellcolor[HTML]{D0EC9F}}0.64 & {\cellcolor[HTML]{DDF2A6}}0.56 & {\cellcolor[HTML]{CFEC9E}}0.65 \\
\bottomrule
Writing script $\rightarrow$ & Lat & Lat & Lat & Lat & Lat & Lat & Lat & Lat & Lat & Lat & Lat & Cyr & Cyr & Cyr & Lat & Grk & Arab & Han & \\
\end{tabular}
}
\caption{\textbf{(RQ1)}  Per-language macro-averaged $F_1$ scores of the selected methods on test data. Abbreviations of writing scripts are as follows: Lat = Latin, Cyr = Cyrillic, Grk = Greek, Arab = Arabic, Han = Hanzi. Bolded values indicate the best method for each test language. Darker shades of green indicate higher scores.} 
\label{multilingualresults}
\end{table*}

\vspace{2mm}
\section{Results}
\label{results}
We present  our experimental results  in Sect. \ref{subsec:multilingual} for the RQ1 task,   in Sect. \ref{subsec:crosslingual} for the   RQ2 tasks,  and in Sect. \ref{subsec:generator-influence} for the RQ3 task.

\subsection{Multilingual Suitability Evaluation}
\label{subsec:multilingual}

Table~\ref{multilingualresults} shows performance results  (macro-averaged  $F_1$) achieved by the methods in our \MMGT{} problem setting based on 18 different languages. 
For reference, a random classifier performance has 0.125  macro $F_1$, due to distinguishing among 8 fully balanced classes. 

At a first glance, we notice that five out of the eight    detectors achieve  macro $F_1 \geq 0.75$. % 
Fine-tuning and contrastive approaches appear to help a lot in adaptability to the multilingual task, with the three best detectors,  Qwen3-4B-Base, mdok and OTBDetector, remarkably showing an $F_1$ score above 0.9 in most cases, across all tested languages. Interestingly, OTBDetector appears to boost generalizability relatively better than mdok and Qwen3-4B-Base if we consider that, despite being $7\times$ smaller than them in parameter size, the $F_1$ score of OTBDetector only reduces by 3\%, which might be due to a sharper decision boundary as determined by the contrastive loss used in OTBDetector. 

As expected, detectors based on a multilingual pretraining (i.e., Qwen3-4B-Base and mdok, OTBDetector, XLM-RoBERTa-large) exhibit stronger multilingual generalization compared to monolingual ones; however, it happens that  English texts are generally difficult to attribute, even for English-only-pretrained methods like RoBERTa. 

We ascribe this behavior to the fact that, since 
English is typically   the best-supported language for most LLMs, the generator outputs might be harder to distinguish because they are more fluent and human-like, yet generators may converge stylistically, and differences between generators become subtle and blurry.
In addition, XLM-RoBERTa-large   performs worse than RoBERTa-large on the English portion of the multilingual test set, which can be explained since   XLM-RoBERTa-large was originally pretrained on tens of languages simultaneously, and hence its capacity is spread across multiple languages, meaning its English representation is less specialized.

Finally, statistical approaches seem to be struggling overall across the results in Table~\ref{multilingualresults}. This is explained since they  are conceived to simply separate human-written from machine-generated texts based on statistical patterns, which may not generalize to attribution. Furthermore, as most of these approaches rely on distributional patterns, their performance collapses in languages where  LLM generators are very proficient---and thus adhere to human-like distributions---or in non-Latin scripts, which present distributional mismatches to Latin ones. Our conjecture is supported by the Binoculars case: it leverages the Falcon 7B model, which was trained mostly on English, German, Spanish, and French, i.e., Latin-script languages. Consequently, its representations are poorly suited to Cyrillic- or Arabic-script inputs, leading to failure in attributing texts in these languages.

\subsection{Cross-lingual Transferability Evaluation}
\label{subsec:crosslingual}

\begin{table*}[!t]
\centering
\setlength{\tabcolsep}{1.0mm}
\scalebox{0.73}{
\begin{tabular}{c|l||c|c|c||c|c|c||c|c|c|c|c||c|c|c||c|c|c|c||c}
\multicolumn{2}{r||}{Lang. family $\rightarrow$} & \multicolumn{3}{c||}{Germanic} & \multicolumn{3}{c||}{Romance} & \multicolumn{5}{c||}{Slavic-Latin} & \multicolumn{3}{c||}{Slavic-Cyrillic} & \multicolumn{4}{c||}{Others} & \\
\midrule
& Method $\downarrow$ & de & en & nl & es & pt & ro & cs & hr & pl & sk & sl & bg & ru & uk & hu & el & ar & zh & all \\
\midrule
\multirow[c]{6}{*}{en} & Qwen3-4B-Base & {\cellcolor[HTML]{F8FCBE}}0.33 & {\cellcolor[HTML]{9FD788}}0.89 & {\cellcolor[HTML]{F8FCC0}}0.32 & {\cellcolor[HTML]{EDF8B2}}0.44 & {\cellcolor[HTML]{E3F4AA}}0.52 & {\cellcolor[HTML]{FAFDCB}}0.27 & {\cellcolor[HTML]{FAFDCC}}0.26 & {\cellcolor[HTML]{FBFDCF}}0.25 & {\cellcolor[HTML]{FCFED3}}0.22 & {\cellcolor[HTML]{FCFED3}}0.22 & {\cellcolor[HTML]{FDFEDD}}0.17 & {\cellcolor[HTML]{FDFED9}}0.20 & {\cellcolor[HTML]{F9FDC7}}0.29 & {\cellcolor[HTML]{FCFED7}}0.20 & {\cellcolor[HTML]{FFFFE5}}{\cellcolor{white}}0.12 & {\cellcolor[HTML]{FFFFE5}}{\cellcolor{white}}0.09 & {\cellcolor[HTML]{FCFED6}}0.21 & {\cellcolor[HTML]{FFFFE5}}{\cellcolor{white}}0.09 & {\cellcolor[HTML]{F9FDC5}}0.30 \\
 & mdok & {\cellcolor[HTML]{E6F5AC}}0.50 & {\cellcolor[HTML]{9CD687}}\bfseries 0.90 & {\cellcolor[HTML]{F4FBB7}}0.39 & {\cellcolor[HTML]{DEF2A7}}\bfseries 0.55 & {\cellcolor[HTML]{DAF0A4}}\bfseries 0.59 & {\cellcolor[HTML]{F8FCBD}}0.34 & {\cellcolor[HTML]{F9FDC2}}0.31 & {\cellcolor[HTML]{FAFDC8}}0.29 & {\cellcolor[HTML]{F8FDC1}}0.32 & {\cellcolor[HTML]{FCFED7}}0.20 & {\cellcolor[HTML]{FCFED4}}0.22 & {\cellcolor[HTML]{FFFFE5}}0.13 & {\cellcolor[HTML]{FCFED4}}0.22 & {\cellcolor[HTML]{FDFEDB}}0.18 & {\cellcolor[HTML]{FAFDCC}}0.26 & {\cellcolor[HTML]{FFFFE5}}{\cellcolor{white}}0.10 & {\cellcolor[HTML]{FFFFE5}}{\cellcolor{white}}0.10 & {\cellcolor[HTML]{FFFFE5}}{\cellcolor{white}}0.12 & {\cellcolor[HTML]{F7FCBA}}0.36 \\
 & OTBDetector & {\cellcolor[HTML]{E4F4AB}}0.51 & {\cellcolor[HTML]{AEDD8E}}0.83 & {\cellcolor[HTML]{EFF9B3}}0.42 & {\cellcolor[HTML]{E9F6AF}}0.47 & {\cellcolor[HTML]{E1F3A9}}0.53 & {\cellcolor[HTML]{EAF7AF}}\bfseries 0.46 & {\cellcolor[HTML]{EFF9B3}}\bfseries 0.42 & {\cellcolor[HTML]{F1FAB5}}\bfseries 0.40 & {\cellcolor[HTML]{EFF9B3}}\bfseries 0.42 & {\cellcolor[HTML]{F7FCBC}}\bfseries 0.35 & {\cellcolor[HTML]{F7FCB9}}\bfseries 0.36 & {\cellcolor[HTML]{F7FCBC}}\bfseries 0.35 & {\cellcolor[HTML]{F4FBB7}}\bfseries 0.39 & {\cellcolor[HTML]{F7FCBA}}\bfseries 0.36 & {\cellcolor[HTML]{F8FCBD}}\bfseries 0.34 & {\cellcolor[HTML]{FAFDC8}}\bfseries 0.29 & {\cellcolor[HTML]{FAFDC9}}\bfseries 0.27 & {\cellcolor[HTML]{FBFDCF}}\bfseries 0.25 & {\cellcolor[HTML]{EEF9B3}}\bfseries 0.43 \\
 & XLM-R-large & {\cellcolor[HTML]{E3F4AA}}\bfseries 0.52 & {\cellcolor[HTML]{DBF1A4}}0.58 & {\cellcolor[HTML]{EEF9B3}}\bfseries 0.43 & {\cellcolor[HTML]{F7FCBA}}0.36 & {\cellcolor[HTML]{EEF9B3}}0.43 & {\cellcolor[HTML]{F6FCB8}}0.37 & {\cellcolor[HTML]{F8FCC0}}0.33 & {\cellcolor[HTML]{F8FCC0}}0.33 & {\cellcolor[HTML]{F7FCB9}}0.37 & {\cellcolor[HTML]{FAFDCB}}0.27 & {\cellcolor[HTML]{F9FDC7}}0.29 & {\cellcolor[HTML]{F9FDC5}}0.30 & {\cellcolor[HTML]{F7FCB9}}0.36 & {\cellcolor[HTML]{F9FDC2}}0.31 & {\cellcolor[HTML]{FAFDCB}}0.27 & {\cellcolor[HTML]{FCFED4}}0.21 & {\cellcolor[HTML]{FCFED6}}0.21 & {\cellcolor[HTML]{FEFFDF}}0.16 & {\cellcolor[HTML]{F6FCB8}}0.37 \\
 & RoBERTa-large & {\cellcolor[HTML]{FFFFE5}}{\cellcolor{white}}0.10 & {\cellcolor[HTML]{CEEB9E}}0.66 & {\cellcolor[HTML]{FFFFE5}}{\cellcolor{white}}0.05 & {\cellcolor[HTML]{FFFFE5}}{\cellcolor{white}}0.11 & {\cellcolor[HTML]{FFFFE5}}{\cellcolor{white}}0.09 & {\cellcolor[HTML]{FFFFE5}}{\cellcolor{white}}0.08 & {\cellcolor[HTML]{FFFFE5}}{\cellcolor{white}}0.10 & {\cellcolor[HTML]{FFFFE5}}{\cellcolor{white}}0.05 & {\cellcolor[HTML]{FFFFE5}}{\cellcolor{white}}0.10 & {\cellcolor[HTML]{FFFFE5}}{\cellcolor{white}}0.08 & {\cellcolor[HTML]{FFFFE5}}{\cellcolor{white}}0.04 & {\cellcolor[HTML]{FFFFE5}}{\cellcolor{white}}0.05 & {\cellcolor[HTML]{FFFFE5}}{\cellcolor{white}}0.05 & {\cellcolor[HTML]{FFFFE5}}{\cellcolor{white}}0.04 & {\cellcolor[HTML]{FFFFE5}}{\cellcolor{white}}0.06 & {\cellcolor[HTML]{FFFFE5}}{\cellcolor{white}}0.05 & {\cellcolor[HTML]{FFFFE5}}{\cellcolor{white}}0.05 & {\cellcolor[HTML]{FFFFE5}}{\cellcolor{white}}0.05 & {\cellcolor[HTML]{FFFFE5}}0.13 \\
 & StatEnsemble & {\cellcolor[HTML]{FCFED4}}0.21 & {\cellcolor[HTML]{E1F3A9}}0.53 & {\cellcolor[HTML]{FDFED9}}0.20 & {\cellcolor[HTML]{FAFDCB}}0.27 & {\cellcolor[HTML]{FCFED3}}0.23 & {\cellcolor[HTML]{FFFFE5}}{\cellcolor{white}}0.10 & {\cellcolor[HTML]{FFFFE5}}{\cellcolor{white}}0.09 & {\cellcolor[HTML]{FFFFE5}}{\cellcolor{white}}0.11 & {\cellcolor[HTML]{FFFFE5}}{\cellcolor{white}}0.10 & {\cellcolor[HTML]{FFFFE5}}{\cellcolor{white}}0.07 & {\cellcolor[HTML]{FFFFE5}}{\cellcolor{white}}0.09 & {\cellcolor[HTML]{FFFFE5}}{\cellcolor{white}}0.08 & {\cellcolor[HTML]{FFFFE5}}0.13 & {\cellcolor[HTML]{FFFFE5}}{\cellcolor{white}}0.02 & {\cellcolor[HTML]{FFFFE5}}{\cellcolor{white}}0.11 & {\cellcolor[HTML]{FFFFE5}}{\cellcolor{white}}0.03 & {\cellcolor[HTML]{FEFFDF}}0.16 & {\cellcolor[HTML]{FDFED9}}0.19 & {\cellcolor[HTML]{FEFFDE}}0.16 \\
\midrule
\multirow[c]{6}{*}{es} & Qwen3-4B-Base & {\cellcolor[HTML]{BEE596}}\bfseries 0.74 & {\cellcolor[HTML]{C7E89A}}\bfseries 0.69 & {\cellcolor[HTML]{CCEA9D}}\bfseries 0.66 & {\cellcolor[HTML]{9DD688}}\bfseries 0.90 & {\cellcolor[HTML]{A4D98A}}\bfseries 0.87 & {\cellcolor[HTML]{CCEA9D}}0.66 & {\cellcolor[HTML]{DDF1A6}}\bfseries 0.57 & {\cellcolor[HTML]{EDF8B2}}0.44 & {\cellcolor[HTML]{CCEA9D}}0.66 & {\cellcolor[HTML]{E5F5AC}}\bfseries 0.50 & {\cellcolor[HTML]{EEF9B3}}0.43 & {\cellcolor[HTML]{F1FAB5}}0.41 & {\cellcolor[HTML]{DDF1A6}}0.57 & {\cellcolor[HTML]{E9F6AF}}0.47 & {\cellcolor[HTML]{EDF8B2}}0.43 & {\cellcolor[HTML]{FCFED3}}0.23 & {\cellcolor[HTML]{F8FDC1}}0.32 & {\cellcolor[HTML]{FFFFE5}}{\cellcolor{white}}0.12 & {\cellcolor[HTML]{DDF2A6}}\bfseries 0.56 \\
 & mdok & {\cellcolor[HTML]{C9E99C}}0.68 & {\cellcolor[HTML]{CCEA9D}}0.66 & {\cellcolor[HTML]{D9F0A3}}0.60 & {\cellcolor[HTML]{9FD788}}0.89 & {\cellcolor[HTML]{ABDC8D}}0.84 & {\cellcolor[HTML]{CFEC9E}}0.65 & {\cellcolor[HTML]{E7F6AD}}0.49 & {\cellcolor[HTML]{E9F6AF}}0.47 & {\cellcolor[HTML]{D3EDA0}}0.62 & {\cellcolor[HTML]{F3FAB6}}0.39 & {\cellcolor[HTML]{EEF9B3}}0.43 & {\cellcolor[HTML]{FAFDC9}}0.28 & {\cellcolor[HTML]{EBF7B0}}0.46 & {\cellcolor[HTML]{F1FAB5}}0.41 & {\cellcolor[HTML]{EDF8B2}}\bfseries 0.43 & {\cellcolor[HTML]{FDFED9}}0.19 & {\cellcolor[HTML]{FCFED6}}0.21 & {\cellcolor[HTML]{FCFED7}}0.20 & {\cellcolor[HTML]{E2F4AA}}0.52 \\
 & OTBDetector & {\cellcolor[HTML]{CFEC9E}}0.65 & {\cellcolor[HTML]{D9F0A3}}0.60 & {\cellcolor[HTML]{D2EDA0}}0.64 & {\cellcolor[HTML]{B6E192}}0.78 & {\cellcolor[HTML]{B3E091}}0.80 & {\cellcolor[HTML]{C8E99B}}\bfseries 0.69 & {\cellcolor[HTML]{E3F4AA}}0.52 & {\cellcolor[HTML]{E6F5AC}}\bfseries 0.50 & {\cellcolor[HTML]{CBEA9C}}\bfseries 0.67 & {\cellcolor[HTML]{EDF8B1}}0.44 & {\cellcolor[HTML]{EDF8B2}}\bfseries 0.44 & {\cellcolor[HTML]{E8F6AE}}\bfseries 0.48 & {\cellcolor[HTML]{DCF1A5}}\bfseries 0.57 & {\cellcolor[HTML]{E3F4AA}}\bfseries 0.52 & {\cellcolor[HTML]{F5FBB8}}0.38 & {\cellcolor[HTML]{F7FCBA}}\bfseries 0.36 & {\cellcolor[HTML]{F2FAB5}}\bfseries 0.40 & {\cellcolor[HTML]{F8FCC0}}\bfseries 0.32 & {\cellcolor[HTML]{DEF2A7}}0.56 \\
 & XLM-R-large & {\cellcolor[HTML]{DCF1A5}}0.57 & {\cellcolor[HTML]{F3FAB6}}0.39 & {\cellcolor[HTML]{E0F3A8}}0.54 & {\cellcolor[HTML]{DBF1A4}}0.58 & {\cellcolor[HTML]{DFF3A8}}0.55 & {\cellcolor[HTML]{F2FAB5}}0.40 & {\cellcolor[HTML]{F3FAB6}}0.39 & {\cellcolor[HTML]{F8FDC1}}0.32 & {\cellcolor[HTML]{F3FAB6}}0.39 & {\cellcolor[HTML]{F8FCBE}}0.33 & {\cellcolor[HTML]{F8FDC1}}0.32 & {\cellcolor[HTML]{F7FCBA}}0.36 & {\cellcolor[HTML]{F5FBB8}}0.37 & {\cellcolor[HTML]{F8FCBE}}0.34 & {\cellcolor[HTML]{F9FDC2}}0.31 & {\cellcolor[HTML]{FAFDCC}}0.26 & {\cellcolor[HTML]{FBFDCE}}0.25 & {\cellcolor[HTML]{FCFED7}}0.21 & {\cellcolor[HTML]{F1FAB5}}0.41 \\
 & RoBERTa-large & {\cellcolor[HTML]{E8F6AE}}0.48 & {\cellcolor[HTML]{FCFED7}}0.20 & {\cellcolor[HTML]{E2F4AA}}0.53 & {\cellcolor[HTML]{DBF1A4}}0.58 & {\cellcolor[HTML]{D9F0A3}}0.60 & {\cellcolor[HTML]{D7EFA2}}0.60 & {\cellcolor[HTML]{F7FCBC}}0.35 & {\cellcolor[HTML]{ECF7B1}}0.45 & {\cellcolor[HTML]{FBFDCE}}0.25 & {\cellcolor[HTML]{FBFED0}}0.24 & {\cellcolor[HTML]{FCFED3}}0.22 & {\cellcolor[HTML]{FFFFE5}}{\cellcolor{white}}0.05 & {\cellcolor[HTML]{FFFFE5}}{\cellcolor{white}}0.04 & {\cellcolor[HTML]{FFFFE5}}{\cellcolor{white}}0.04 & {\cellcolor[HTML]{FDFED9}}0.19 & {\cellcolor[HTML]{FFFFE5}}{\cellcolor{white}}0.06 & {\cellcolor[HTML]{FFFFE5}}{\cellcolor{white}}0.07 & {\cellcolor[HTML]{FFFFE5}}{\cellcolor{white}}0.07 & {\cellcolor[HTML]{F8FDC1}}0.32 \\
 & StatEnsemble & {\cellcolor[HTML]{F8FDC1}}0.32 & {\cellcolor[HTML]{F7FCBA}}0.36 & {\cellcolor[HTML]{F7FCBA}}0.36 & {\cellcolor[HTML]{E6F5AC}}0.49 & {\cellcolor[HTML]{ECF7B1}}0.45 & {\cellcolor[HTML]{FAFDCB}}0.27 & {\cellcolor[HTML]{FBFED0}}0.24 & {\cellcolor[HTML]{FDFED9}}0.20 & {\cellcolor[HTML]{FBFDCF}}0.25 & {\cellcolor[HTML]{FEFFDF}}0.16 & {\cellcolor[HTML]{FFFFE5}}{\cellcolor{white}}0.11 & {\cellcolor[HTML]{F9FDC7}}0.29 & {\cellcolor[HTML]{F9FDC7}}0.29 & {\cellcolor[HTML]{FFFFE5}}{\cellcolor{white}}0.12 & {\cellcolor[HTML]{FAFDC8}}0.28 & {\cellcolor[HTML]{FBFED0}}0.24 & {\cellcolor[HTML]{FBFDCE}}0.26 & {\cellcolor[HTML]{FDFED9}}0.20 & {\cellcolor[HTML]{FAFDC8}}0.28 \\
\midrule
\multirow[c]{6}{*}{ru} & Qwen3-4B-Base & {\cellcolor[HTML]{D0EC9F}}0.64 & {\cellcolor[HTML]{F1FAB5}}0.40 & {\cellcolor[HTML]{CFEC9E}}\bfseries 0.65 & {\cellcolor[HTML]{D7EFA2}}0.61 & {\cellcolor[HTML]{D0EC9F}}0.64 & {\cellcolor[HTML]{C3E698}}0.72 & {\cellcolor[HTML]{B8E293}}0.77 & {\cellcolor[HTML]{BEE596}}0.73 & {\cellcolor[HTML]{BEE596}}0.74 & {\cellcolor[HTML]{B3E091}}\bfseries 0.79 & {\cellcolor[HTML]{BAE394}}0.76 & {\cellcolor[HTML]{B2DF90}}0.80 & {\cellcolor[HTML]{A4D98A}}0.87 & {\cellcolor[HTML]{B8E293}}0.77 & {\cellcolor[HTML]{E6F5AC}}0.50 & {\cellcolor[HTML]{EDF8B2}}0.44 & {\cellcolor[HTML]{E4F4AB}}\bfseries 0.51 & {\cellcolor[HTML]{FAFDCC}}0.26 & {\cellcolor[HTML]{CBEA9C}}0.67 \\
 & mdok & {\cellcolor[HTML]{C0E597}}\bfseries 0.73 & {\cellcolor[HTML]{E6F5AC}}\bfseries 0.49 & {\cellcolor[HTML]{CFEC9E}}0.65 & {\cellcolor[HTML]{D2EDA0}}\bfseries 0.63 & {\cellcolor[HTML]{C1E698}}\bfseries 0.72 & {\cellcolor[HTML]{C3E698}}\bfseries 0.72 & {\cellcolor[HTML]{B2DF90}}\bfseries 0.80 & {\cellcolor[HTML]{BCE395}}\bfseries 0.75 & {\cellcolor[HTML]{B8E293}}\bfseries 0.77 & {\cellcolor[HTML]{BEE596}}0.74 & {\cellcolor[HTML]{B5E092}}\bfseries 0.79 & {\cellcolor[HTML]{B2DF90}}\bfseries 0.80 & {\cellcolor[HTML]{A2D88A}}\bfseries 0.88 & {\cellcolor[HTML]{B2DF90}}0.80 & {\cellcolor[HTML]{C5E89A}}\bfseries 0.70 & {\cellcolor[HTML]{F4FBB7}}0.38 & {\cellcolor[HTML]{EFF9B3}}0.42 & {\cellcolor[HTML]{F8FDC1}}0.32 & {\cellcolor[HTML]{C8E99B}}\bfseries 0.68 \\
 & OTBDetector & {\cellcolor[HTML]{D2EDA0}}0.63 & {\cellcolor[HTML]{EFF9B3}}0.42 & {\cellcolor[HTML]{E1F3A9}}0.53 & {\cellcolor[HTML]{EEF9B3}}0.43 & {\cellcolor[HTML]{EAF7AF}}0.47 & {\cellcolor[HTML]{DFF3A8}}0.55 & {\cellcolor[HTML]{B3E091}}0.80 & {\cellcolor[HTML]{C4E799}}0.71 & {\cellcolor[HTML]{BEE596}}0.73 & {\cellcolor[HTML]{C0E597}}0.73 & {\cellcolor[HTML]{C3E698}}0.72 & {\cellcolor[HTML]{B8E293}}0.78 & {\cellcolor[HTML]{B2DF90}}0.80 & {\cellcolor[HTML]{ACDD8E}}\bfseries 0.83 & {\cellcolor[HTML]{CEEB9E}}0.65 & {\cellcolor[HTML]{E7F6AD}}\bfseries 0.49 & {\cellcolor[HTML]{EAF7AF}}0.46 & {\cellcolor[HTML]{ECF7B1}}\bfseries 0.45 & {\cellcolor[HTML]{D0EC9F}}0.64 \\
 & XLM-R-large & {\cellcolor[HTML]{EEF9B3}}0.43 & {\cellcolor[HTML]{FBFED2}}0.23 & {\cellcolor[HTML]{F9FDC4}}0.30 & {\cellcolor[HTML]{F9FDC4}}0.30 & {\cellcolor[HTML]{F9FDC4}}0.30 & {\cellcolor[HTML]{EDF8B1}}0.44 & {\cellcolor[HTML]{C0E597}}0.73 & {\cellcolor[HTML]{DAF0A4}}0.59 & {\cellcolor[HTML]{D5EEA1}}0.62 & {\cellcolor[HTML]{D0EC9F}}0.65 & {\cellcolor[HTML]{CBEA9C}}0.67 & {\cellcolor[HTML]{CBEA9C}}0.67 & {\cellcolor[HTML]{D2EDA0}}0.63 & {\cellcolor[HTML]{C7E89A}}0.69 & {\cellcolor[HTML]{D0EC9F}}0.64 & {\cellcolor[HTML]{F3FAB6}}0.40 & {\cellcolor[HTML]{EEF9B3}}0.43 & {\cellcolor[HTML]{F7FCB9}}0.37 & {\cellcolor[HTML]{E2F4AA}}0.53 \\
 & RoBERTa-large & {\cellcolor[HTML]{FFFFE5}}{\cellcolor{white}}0.07 & {\cellcolor[HTML]{FFFFE5}}{\cellcolor{white}}0.05 & {\cellcolor[HTML]{FFFFE5}}{\cellcolor{white}}0.06 & {\cellcolor[HTML]{FFFFE5}}{\cellcolor{white}}0.07 & {\cellcolor[HTML]{FFFFE5}}{\cellcolor{white}}0.07 & {\cellcolor[HTML]{FFFFE5}}{\cellcolor{white}}0.08 & {\cellcolor[HTML]{FFFFE5}}{\cellcolor{white}}0.06 & {\cellcolor[HTML]{FFFFE5}}{\cellcolor{white}}0.09 & {\cellcolor[HTML]{FFFFE5}}{\cellcolor{white}}0.07 & {\cellcolor[HTML]{FFFFE5}}{\cellcolor{white}}0.09 & {\cellcolor[HTML]{FFFFE5}}{\cellcolor{white}}0.08 & {\cellcolor[HTML]{F4FBB7}}0.38 & {\cellcolor[HTML]{EDF8B2}}0.43 & {\cellcolor[HTML]{F5FBB8}}0.38 & {\cellcolor[HTML]{FFFFE5}}{\cellcolor{white}}0.06 & {\cellcolor[HTML]{FCFED3}}0.22 & {\cellcolor[HTML]{FEFFE1}}0.15 & {\cellcolor[HTML]{FFFFE5}}{\cellcolor{white}}0.09 & {\cellcolor[HTML]{FDFEDD}}0.17 \\
 & StatEnsemble & {\cellcolor[HTML]{F9FDC5}}0.29 & {\cellcolor[HTML]{FFFFE4}}0.13 & {\cellcolor[HTML]{F9FDC4}}0.30 & {\cellcolor[HTML]{FCFED3}}0.23 & {\cellcolor[HTML]{FBFDCE}}0.26 & {\cellcolor[HTML]{FAFDC9}}0.27 & {\cellcolor[HTML]{FAFDCC}}0.26 & {\cellcolor[HTML]{FBFDCE}}0.26 & {\cellcolor[HTML]{F0F9B4}}0.41 & {\cellcolor[HTML]{FCFED3}}0.23 & {\cellcolor[HTML]{FEFFDE}}0.16 & {\cellcolor[HTML]{E5F5AC}}0.50 & {\cellcolor[HTML]{E5F5AC}}0.50 & {\cellcolor[HTML]{F9FDC7}}0.29 & {\cellcolor[HTML]{F0F9B4}}0.42 & {\cellcolor[HTML]{FAFDC8}}0.29 & {\cellcolor[HTML]{FAFDCC}}0.26 & {\cellcolor[HTML]{FBFED0}}0.24 & {\cellcolor[HTML]{F8FCC0}}0.32 \\
\midrule
\multirow[c]{6}{*}{en-es-ru} & Qwen3-4B-Base & {\cellcolor[HTML]{B2DF90}}\bfseries 0.81 & {\cellcolor[HTML]{A9DB8C}}0.84 & {\cellcolor[HTML]{B5E092}}0.79 & {\cellcolor[HTML]{B1DF90}}0.81 & {\cellcolor[HTML]{ABDC8D}}0.84 & {\cellcolor[HTML]{ABDC8D}}\bfseries 0.84 & {\cellcolor[HTML]{B2DF90}}\bfseries 0.80 & {\cellcolor[HTML]{C0E597}}0.73 & {\cellcolor[HTML]{B8E293}}0.77 & {\cellcolor[HTML]{BEE596}}\bfseries 0.74 & {\cellcolor[HTML]{C0E597}}\bfseries 0.73 & {\cellcolor[HTML]{B8E293}}\bfseries 0.77 & {\cellcolor[HTML]{B3E091}}0.80 & {\cellcolor[HTML]{C1E698}}0.72 & {\cellcolor[HTML]{CFEC9E}}\bfseries 0.65 & {\cellcolor[HTML]{EFF9B3}}\bfseries 0.42 & {\cellcolor[HTML]{EEF9B3}}0.43 & {\cellcolor[HTML]{F9FDC4}}0.31 & {\cellcolor[HTML]{C1E698}}0.72 \\
 & mdok & {\cellcolor[HTML]{B9E294}}0.77 & {\cellcolor[HTML]{A1D889}}\bfseries 0.88 & {\cellcolor[HTML]{B5E092}}\bfseries 0.79 & {\cellcolor[HTML]{A4D98A}}\bfseries 0.86 & {\cellcolor[HTML]{A6DA8B}}\bfseries 0.86 & {\cellcolor[HTML]{B5E092}}0.78 & {\cellcolor[HTML]{BCE395}}0.75 & {\cellcolor[HTML]{BEE596}}\bfseries 0.73 & {\cellcolor[HTML]{B2DF90}}\bfseries 0.80 & {\cellcolor[HTML]{CBEA9C}}0.67 & {\cellcolor[HTML]{C1E698}}0.72 & {\cellcolor[HTML]{B9E294}}0.76 & {\cellcolor[HTML]{A7DB8C}}\bfseries 0.85 & {\cellcolor[HTML]{B6E192}}\bfseries 0.78 & {\cellcolor[HTML]{D0EC9F}}0.64 & {\cellcolor[HTML]{F3FAB6}}0.39 & {\cellcolor[HTML]{EBF7B0}}\bfseries 0.46 & {\cellcolor[HTML]{F7FCBC}}\bfseries 0.35 & {\cellcolor[HTML]{C1E698}}\bfseries 0.72 \\
 & OTBDetector & {\cellcolor[HTML]{C7E89A}}0.69 & {\cellcolor[HTML]{C5E89A}}0.70 & {\cellcolor[HTML]{C8E99B}}0.69 & {\cellcolor[HTML]{BEE596}}0.74 & {\cellcolor[HTML]{BDE496}}0.74 & {\cellcolor[HTML]{C5E89A}}0.70 & {\cellcolor[HTML]{C8E99B}}0.69 & {\cellcolor[HTML]{D2EDA0}}0.64 & {\cellcolor[HTML]{C4E799}}0.71 & {\cellcolor[HTML]{DDF2A6}}0.56 & {\cellcolor[HTML]{DAF0A4}}0.59 & {\cellcolor[HTML]{CEEB9E}}0.66 & {\cellcolor[HTML]{BAE394}}0.76 & {\cellcolor[HTML]{BAE394}}0.76 & {\cellcolor[HTML]{E4F4AB}}0.51 & {\cellcolor[HTML]{EFF9B3}}0.42 & {\cellcolor[HTML]{EFF9B3}}0.42 & {\cellcolor[HTML]{F8FCBD}}0.34 & {\cellcolor[HTML]{D0EC9F}}0.64 \\
 & XLM-R-large & {\cellcolor[HTML]{DDF1A6}}0.57 & {\cellcolor[HTML]{EDF8B2}}0.44 & {\cellcolor[HTML]{EAF7AF}}0.47 & {\cellcolor[HTML]{E8F6AE}}0.48 & {\cellcolor[HTML]{EAF7AF}}0.46 & {\cellcolor[HTML]{E2F4AA}}0.53 & {\cellcolor[HTML]{D0EC9F}}0.64 & {\cellcolor[HTML]{DAF0A4}}0.59 & {\cellcolor[HTML]{D6EFA2}}0.61 & {\cellcolor[HTML]{DDF2A6}}0.56 & {\cellcolor[HTML]{D2EDA0}}0.63 & {\cellcolor[HTML]{DAF0A4}}0.59 & {\cellcolor[HTML]{DDF2A6}}0.56 & {\cellcolor[HTML]{DDF2A6}}0.56 & {\cellcolor[HTML]{D9F0A3}}0.60 & {\cellcolor[HTML]{F7FCBA}}0.36 & {\cellcolor[HTML]{F0F9B4}}0.41 & {\cellcolor[HTML]{F8FCBE}}0.34 & {\cellcolor[HTML]{E1F3A9}}0.53 \\
 & RoBERTa-large & {\cellcolor[HTML]{EBF7B0}}0.46 & {\cellcolor[HTML]{DCF1A5}}0.57 & {\cellcolor[HTML]{E0F3A8}}0.54 & {\cellcolor[HTML]{E8F6AE}}0.48 & {\cellcolor[HTML]{DEF2A7}}0.56 & {\cellcolor[HTML]{DFF3A8}}0.55 & {\cellcolor[HTML]{E9F6AF}}0.47 & {\cellcolor[HTML]{E6F5AC}}0.49 & {\cellcolor[HTML]{F8FDC1}}0.32 & {\cellcolor[HTML]{F9FDC7}}0.29 & {\cellcolor[HTML]{FCFED3}}0.22 & {\cellcolor[HTML]{F6FCB8}}0.37 & {\cellcolor[HTML]{F2FAB5}}0.40 & {\cellcolor[HTML]{F3FAB6}}0.39 & {\cellcolor[HTML]{FBFED2}}0.23 & {\cellcolor[HTML]{FBFDCF}}0.25 & {\cellcolor[HTML]{FDFEDB}}0.18 & {\cellcolor[HTML]{FEFFE1}}0.15 & {\cellcolor[HTML]{F2FAB5}}0.40 \\
 & StatEnsemble & {\cellcolor[HTML]{F7FCBC}}0.35 & {\cellcolor[HTML]{F1FAB5}}0.41 & {\cellcolor[HTML]{F0F9B4}}0.42 & {\cellcolor[HTML]{EDF8B2}}0.43 & {\cellcolor[HTML]{EFF9B3}}0.42 & {\cellcolor[HTML]{F9FDC5}}0.30 & {\cellcolor[HTML]{F9FDC4}}0.30 & {\cellcolor[HTML]{FAFDC8}}0.28 & {\cellcolor[HTML]{F6FCB8}}0.37 & {\cellcolor[HTML]{FCFED3}}0.23 & {\cellcolor[HTML]{FDFEDB}}0.18 & {\cellcolor[HTML]{F2FAB5}}0.40 & {\cellcolor[HTML]{F0F9B4}}0.42 & {\cellcolor[HTML]{FCFED6}}0.21 & {\cellcolor[HTML]{F6FCB8}}0.37 & {\cellcolor[HTML]{FAFDCB}}0.27 & {\cellcolor[HTML]{FAFDC9}}0.28 & {\cellcolor[HTML]{FCFED4}}0.22 & {\cellcolor[HTML]{F8FCBD}}0.34 \\
\bottomrule
\multicolumn{2}{r||}{Writing script $\rightarrow$} & Lat & Lat & Lat & Lat & Lat & Lat & Lat & Lat & Lat & Lat & Lat & Cyr & Cyr & Cyr & Lat & Grk & Arab & Han & \\
\end{tabular}
}
\caption{\textbf{(RQ2)} Per-language cross-lingual   macro-averaged  $F_1$ scores of the selected methods on test data. Writing scripts are as follows: Lat = Latin, Cyr = Cyrillic, Grk = Greek, Arab = Arabic, Han = Hanzi. Bolded values indicate the best method for each training-language and test-language pair. Darker shades of green indicate higher scores.
}
\label{crosslingualresults}
\end{table*}

\noindent
\textbf{Language-level performance.\ } 
Table \ref{crosslingualresults} reveals several key findings at a language-level, which are summarized as follows. (We hereinafter leave Fast-DetectGPT and Binoculars  out of evaluation given their poor performance as shown in  Table~\ref{multilingualresults}).  
 
 Training on Russian (alone or in combination with others) has a significantly greater impact than other languages on the cross-lingual transferability, 
with +0.25 vs. English and  +0.12 vs. Spanish    in terms of overall best results; moreover, the observed benefit from training on Russian extends also to  languages of a different family, especially non-Latin languages.  
By contrast, English appears to be the least generalizable, even among intra-script languages. This may be due to the simplicity of English tokenization and morphological structure, which fails to capture  well to languages with richer morphological or syntactic complexity.

Focusing on the performance of the three best methods (i.e., Qwen3-4B-Base, mdok and OTBDetector) on the results   corresponding to English, Spanish, and Russian, respectively,  a multilingual model from Table~\ref{multilingualresults} appears to be preferable to a monolingual model trained on language $L$ if the goal is to maximize the prediction performance on $L$ only. 
At first glance, this might be seen as   counterintuitive, since the inclusion of multiple languages in the training set could be expected to dilute language-specific patterns for $L$. However, the exposure to diverse linguistic structures may in fact enhance the model’s generalization ability, even on individual languages. 
Nonetheless, the above remarks should be taken with a grain of salt, as differences in the number of training samples per language may introduce bias into the comparison.

 \vspace{1.5mm}
 \noindent
\textbf{Language-family-level performance.\ } 
Aggregating results from Table \ref{crosslingualresults} by language family (cf. Table \ref{crosslingualresultsperfamily} in the Appendix~\ref{app:per-lang-family}) reveals the beneficial effect of Russian on cross-lingual transferability. Notably, training on Russian alone yields optimal performance in six out of eight test families, i.e., all except Germanic and Romance. For these two families, combining Russian with English and Spanish is essential to maximize performance.

\begin{figure*}[!t]
\centering
\includegraphics[width=\textwidth]{confusion_matrix_mdok_models_internal_external.png} \\
\includegraphics[width=\textwidth]{confusion_matrix_otb_models_internal_external.png} \\
\caption{ 
\textbf{(RQ3)}  Confusion matrices (row-wise percentages) for two of the best-performing approaches, i.e., mdok (top) and OTBDetector (bottom), by varying the training data. \textit{Internal} and \textit{External} here indicate that the method has been evaluated on the same language as training and on all but the training language, respectively. Numbers in the diagonal indicate the percentage of correct predictions.  
LLM generators are referred to using the first letter, namely M = Mistral, O = OPT, E = Eagle, V = Vicuna, L = Llama2, A = Aya, G = GPT-3.5, and H = human.
}
\label{fig:conf_matrices}
\vspace{-1mm}
\end{figure*}

\vspace{1.5mm}
\noindent \textbf{Why Russian languages support better cross-lingual transferability.\ }
We attribute the stronger cross-lingual transferability observed with Russian to a number of syntactic and morphological properties of the language~\cite{wals}. Russian is rich in morphology, with a high inflectional structure, where grammatical roles (e.g., subject, object, verb) are encoded via an extensive use of word endings that allow words to convey a wide range of meanings within a sentence~\cite{wals-49,wals-22}. This contrasts with English, where discourse construction typically relies on fixed syntax and a simpler morphology.
Consequently, its morphological richness may encourage models trained on Russian to capture deeper linguistic signals that transfer more robustly across languages, whereas models trained on English might learn more superficial token-level rules that do not generalize well. This observation is further supported by experiments in Table~\ref{crosslingualresultsslavic}, referring the results of the selected detection methods fine-tuned using the other Slavic languages. They show a similar superiority as the Russian language.

\subsection{Influence of LLM Generators  on \textsf{ML-MGT} and \textsf{CL-MGT}}
\label{subsec:generator-influence}
We analyze the influence that the various   LLM generators have on multilingual authorship attribution  and its cross-lingual transferability  by examining the error patterns of mdok and OTBDetector, which here are selected as they have shown the best trade-off between efficiency and generalizability.  

\vspace{1.5mm}
\noindent\textbf{\MMGT{} Patterns.\ }
Both mdok and OTBDetector exhibit very high 
attribution-performance when trained and evaluated on the full set of available languages,  confirming the remarkable results from Table~\ref{multilingualresults}. Their confusion matrices suggest that those detectors can effectively learn the stylistic footprint of each LLM generator and generalize attribution across languages. 
A closer look at the per-generator behavior (cf. Table~\ref{multilingualresults_generator} in Appendix~\ref{app:per-generator}) reveals that the detectors struggle slightly more in attributing models like Mistral, Eagle, and Vicuna, while excelling in the Aya, GPT-3.5, human, and OPT classes. These differences are also language-dependent, with  Cyrillic languages, Hungarian, Chinese, Czech, and even English showing higher error rates. % 

\vspace{1.5mm}
\noindent\textbf{\CLMGT{} Patterns.\ } 
Figure~\ref{fig:conf_matrices} provides the confusion matrices of the two of the best detectors based on the language selection for our RQ2 (i.e., en, ru, en-es-ru). Here we distinguish between an \textit{Internal} setting, where   training and test languages are the same, and an \textit{External} setting, where the test languages are missing in the training set. 
In the former case,  both mdok and OTBDetector perform well across the three language-group scenarios; however, under the External setting,  performance   tends to worsen, with increasing confusion among architecturally similar models. 

\vspace{1.5mm}
\noindent\textbf{Error Trends by Generator.\ }
Llama2-70B and Vicuna-13B appear to be relatively difficult to attribute, especially in the English-based External setting, which  might be due to the shared underlying architecture among these models---Vicuna is in fact a further-fine-tuning of Llama. 
Interestingly, human-written texts are among the easiest to attribute, suggesting that despite the fluency LLMs have in producing multilingual texts, distinct human-specific patterns remain detectable. 
Finally, OPT-30B and Eagle-7B emerge as the ``catch-all'' classes for English and Russian, respectively, in the External setting, as a recurring pattern for both mdok and OTBDetector involves overpredicting those LLMs. We ascribe this to the tendency of the two LLMs to generate   texts with fewer stylistic variations, thus becoming the most predictable  classes when the detector is uncertain---especially under the \CLMGT{} problem.

\section{Conclusions}
\label{conclusions}
\vspace{-1mm}
Despite the growing multilingual usage of LLMs, current efforts in authorship attribution of MGT remain largely confined to monolingual contexts, particularly English. In this work, we filled this gap by formally defining and exploring the problems of multilingual and cross-lingual authorship attribution in MGT.
We   evaluated the performance of established authorship attribution approaches in the multilingual setting, as well as their ability to generalize across languages. Our experiments, covering 18 languages and 8 author classes (7 LLMs and a human class), demonstrate that while some existing methods can be adapted to multilingual authorship attribution, their effectiveness varies widely, highlighting the challenge of cross-lingual transferability and the need for further development in the field for real-world multilingual usage.  

\vspace{1mm}
\noindent
 Code and data resources associated with this work are available at 
 \url{https://github.com/MLNTeam-Unical/Multilingual-MGT-AA}.

\vspace{1mm}
\noindent 
\textbf{Future work.\ } 
Although the chosen news-domain offers a significant testbed due to its linguistic and topical diversity, our investigation scope should be extended. While we have already obtained preliminary results regarding social contents, as discussed in  Appendix~\ref{app:multisocial},  further analysis on other domains  is left for future work. 
Additionally, we would like to expand our analysis and findings geographically by considering more (low-resource) languages.  Also, we aim to investigate the impact of adversarial attacks on the accuracy and cross-lingual transferability of multilingual MGT attribution approaches.

\vspace{2mm}
\section*{Limitations}

\noindent \textbf{Language Coverage.\ } Our study considers 18 languages from the \multitude{} dataset. While being representative of a broad range of language families and scripts, it does not fully represent the linguistic diversity encountered in real-world settings. For this reason, we are committed to investigating multilingual authorship attribution in more (low-resource) languages.

\vspace{1mm}
\noindent \textbf{Domain Coverage.\ } Our study deliberately focuses on the news domain, as it represents one of the most challenging settings due to high variability in topics, styles, and linguistic registers across languages. While these aspects make attribution intrinsically harder and provide a rigorous testbed for real-world applicability, we acknowledge that attribution performance may differ in other domains (cf. Appendix~\ref{app:multisocial} for insights on social content) and our findings may not strongly generalize beyond news-style content. Expanding our investigations to a broader set of domains is part of future work.

\vspace{1mm}
\noindent \textbf{LLM-generators Coverage.\ }
Our study considers a fixed and controlled set of seven LLM generators and one human-authored class. While these models cover a broad range of architectures and sizes, we acknowledge that the rapidly evolving landscape of NLP might introduce advancements that may degrade attribution methods when faced with unseen or future models, e.g., different data distributions or new decoding strategies.

\vspace{1mm}
\noindent \textbf{Adversarial Attacks.\ }
Our study assumes clean, well-formed text and does not account for adversarial manipulations (e.g., paraphrasing, prompting, obfuscation) aimed at challenging detection and attribution systems. Addressing these scenarios remains an important direction for future work.

\section*{Ethical Considerations}

\noindent\textbf{Broader Impact.\ }
Authorship Attribution systems can be misused or overtrusted, with unintended effects on society. For this reason, we urge all stakeholders to use such methodologies responsibly, keeping in account potential biases across languages (e.g., higher false positives for low-resource languages/scripts), and ensuring humans-in-the-loop while using such tools for decision making.

\section*{Acknowledgments}
This work was partially supported by the European Union under the Horizon Europe project AI-CODE, GA No. \href{https://cordis.europa.eu/project/id/101135437}{101135437}; by the EU NextGenerationEU through the Recovery and Resilience Plan for Slovakia under the project No. 09I01-03-V04-00059; by the project ``Future Artificial Intelligence Research (FAIR)'' spoke 9 (H23C22000860006), and the project SERICS (PE00000014). 

We acknowledge the EuroHPC Joint Undertaking for awarding us access to Leonardo at CINECA, Italy.

\appendix

\clearpage

\section{Computational Resources}
For fine-tuning and inference of authorship attribution  methods (a single run for each version of fine-tuned authorship attribution  method), as well as for hyperparameters optimization, we used a machine allocated with 8 CPU cores (Intel Xeon Platinum 8358 CPU, 2.6 GHz), 128GB RAM, and 1× A100 64GB GPU, cumulatively consuming approximately 500 GPU-hours.  

\section{Multilingual MGT Datasets}
\label{app:ml-datasets}
In Table~\ref{mutlilingualdatasets}, we summarize the basic statistics about generators, languages, and domains of datasets that can be regarded as potentially useful for multilingual authorship attribution.

CUDRT and RAID-extra cover 3 or fewer languages, thus being not well-suited to our cross-lingual study. M4GT-Bench is a composition of multiple datasets covering various domains, which might introduce a bias in the results due to the different nature of the domains.   

To the best of our knowledge, the \multitude{} collection is the only one containing a relevant set of generators, text-generation settings, and domains for each language, enabling a proper cross-lingual transferability evaluation---especially in its latest version.  

It should be noted that MultiSocial also offers broad coverage in terms of both languages and generators. However, it contains an uneven number of samples across languages, and the data are drawn from multiple social media platforms. This heterogeneity may introduce inconsistencies in writing style and topic distribution. 
These considerations motivated our choice of \multitude{} as the target dataset for our study. Nonetheless, as discussed in Appendix \ref{app:multisocial}, we additionally leverage MultiSocial for a preliminary investigation under \textit{out-of-domain conditions}, in order to assess the robustness and generalizability of authorship attribution methods beyond the in-domain (i.e., \multitude{}) setting.

\begin{table*}[!h]
\centering
\scalebox{1}{
\small
\begin{tabular}{l|c|c|c|c}
\toprule
Dataset & Reference & \#Generators & \#Languages & \#Domains\\
\midrule
CUDRT & \cite{tao2024reliabledetectionllmgeneratedtexts} & 5 & 2 & 6 \\
M4GT-Bench & \cite{wang-etal-2024-m4gt} & 8 & 9 & 6 \\
MULTITuDE\_v1 & \cite{macko-etal-2023-multitude} & 9 & 11 & 1 \\
MULTITuDE\_v3 & \cite{macko_2025_15519413} & 8 & 21 & 1 \\
MultiSocial & \cite{macko-etal-2025-multisocial} & 8 & 22 & 1 \\
RAID-extra & \cite{dugan-etal-2024-raid} & 11 & 3 & 8 \\
\bottomrule
\end{tabular}
}
\caption{Overview of existing resources for multilingual machine-generated text detection.  
}
\label{mutlilingualdatasets}
\end{table*}

\section{Out-of-Domain Generalization}
\label{app:multisocial}
To evaluate the generalization of the authorship attribution methods reported in Table~\ref{multilingualresults} to unseen domains, we evaluated them on the MultiSocial data. % 
Indeed, not only it covers a different domain (i.e., social), but it also contains 4 additional languages besides the 18 languages we used for training. 
Results, which are shown in Table \ref{tab:crossdomainresults}, highlight that OTBDetector and XLM-RoBERTa-large achieve better average macro $F_1$ than the other methods, although, as expected, overall performance remains substantially lower across all approaches w.r.t. the in-domain setting.

\begin{table*}[!h]
\centering
\setlength{\tabcolsep}{1.0mm} 
\scalebox{0.65}{
\begin{tabular}{l||c|c|c||c|c|c|c||c|c|c|c|c||c|c|c||c|c|c|c|c|c|c||c}
Lang. family $\rightarrow$ & \multicolumn{3}{c||}{Germanic} & \multicolumn{4}{c||}{Romance} & \multicolumn{5}{c||}{Slavic-Latin} & \multicolumn{3}{c||}{Slavic-Cyrillic} & \multicolumn{7}{c||}{Others} & \\
Method $\downarrow$ & de & en & nl & ca* & es & pt & ro & cs & hr & pl & sk & sl & bg & ru & uk & ga* & gd* & et* & hu & el & ar & zh & all \\
\midrule
OTBDetector & \bfseries {\cellcolor[HTML]{F7FCB9}}0.36 & \bfseries {\cellcolor[HTML]{F8FDC1}}0.32 & {\cellcolor[HTML]{FDFEDB}}0.18 & \bfseries {\cellcolor[HTML]{FBFDCE}}0.25 & \bfseries {\cellcolor[HTML]{FCFED6}}0.21 & \bfseries {\cellcolor[HTML]{F9FDC7}}0.29 & \bfseries {\cellcolor[HTML]{FDFEDA}}0.19 & \bfseries {\cellcolor[HTML]{FCFED6}}0.21 & {\cellcolor[HTML]{FEFFDF}}0.16 & {\cellcolor[HTML]{FEFFDF}}0.16 & {\cellcolor[HTML]{FEFFE1}}0.15 & {\cellcolor[HTML]{FBFED2}}0.23 & {\cellcolor[HTML]{FBFDCE}}0.25 & {\cellcolor[HTML]{FBFED0}}0.24 & {\cellcolor[HTML]{FCFED4}}0.22 & \bfseries {\cellcolor[HTML]{FBFED2}}0.23 & \bfseries {\cellcolor[HTML]{FBFED2}}0.23 & \bfseries {\cellcolor[HTML]{FAFDC8}}0.28 & \bfseries {\cellcolor[HTML]{F8FCC0}}0.33 & \bfseries {\cellcolor[HTML]{F7FCBC}}0.35 & {\cellcolor[HTML]{F9FDC7}}0.29 & {\cellcolor[HTML]{F7FCB9}}0.36 & \bfseries {\cellcolor[HTML]{FBFED0}}0.24 \\
XLM-R-large & {\cellcolor[HTML]{FDFED9}}0.20 & {\cellcolor[HTML]{FEFFE2}}0.14 & \bfseries {\cellcolor[HTML]{FCFED7}}0.21 & {\cellcolor[HTML]{FDFEDD}}0.17 & {\cellcolor[HTML]{FDFEDD}}0.17 & {\cellcolor[HTML]{FDFEDD}}0.17 & {\cellcolor[HTML]{FDFEDD}}0.17 & {\cellcolor[HTML]{FDFED9}}0.19 & \bfseries {\cellcolor[HTML]{FDFEDD}}0.18 & \bfseries {\cellcolor[HTML]{FDFEDB}}0.18 & \bfseries {\cellcolor[HTML]{FAFDC8}}0.28 & \bfseries {\cellcolor[HTML]{FAFDC8}}0.28 & \bfseries {\cellcolor[HTML]{FAFDC8}}0.28 & \bfseries {\cellcolor[HTML]{FAFDCC}}0.26 & \bfseries {\cellcolor[HTML]{FAFDCB}}0.27 & {\cellcolor[HTML]{FEFFE1}}0.15 & {\cellcolor[HTML]{FEFFE1}}0.15 & {\cellcolor[HTML]{FFFFE4}}0.14 & {\cellcolor[HTML]{FDFED9}}0.19 & {\cellcolor[HTML]{FBFDCE}}0.26 & \bfseries {\cellcolor[HTML]{F8FCBD}}0.34 & {\cellcolor[HTML]{F7FCBC}}0.35 & {\cellcolor[HTML]{FCFED6}}0.21 \\
mdok & {\cellcolor[HTML]{FDFEDB}}0.18 & {\cellcolor[HTML]{FDFEDB}}0.18 & {\cellcolor[HTML]{FDFEDB}}0.18 & {\cellcolor[HTML]{FEFFDE}}0.16 & {\cellcolor[HTML]{FDFEDD}}0.17 & {\cellcolor[HTML]{FDFEDB}}0.18 & {\cellcolor[HTML]{FEFFDE}}0.16 & {\cellcolor[HTML]{FDFEDA}}0.19 & {\cellcolor[HTML]{FEFFDE}}0.17 & {\cellcolor[HTML]{FEFFDF}}0.16 & {\cellcolor[HTML]{FBFED2}}0.23 & {\cellcolor[HTML]{FBFED2}}0.23 & {\cellcolor[HTML]{FBFDCF}}0.24 & {\cellcolor[HTML]{FBFDCE}}0.25 & {\cellcolor[HTML]{FCFED4}}0.22 & {\cellcolor[HTML]{FFFFE5}}{\cellcolor{white}}0.09 & {\cellcolor[HTML]{FFFFE5}}{\cellcolor{white}}0.10 & {\cellcolor[HTML]{FFFFE5}}{\cellcolor{white}}0.11 & {\cellcolor[HTML]{FDFEDA}}0.19 & {\cellcolor[HTML]{FBFDCF}}0.24 & {\cellcolor[HTML]{F9FDC5}}0.30 & {\cellcolor[HTML]{F2FAB5}}0.40 & {\cellcolor[HTML]{FDFED9}}0.20 \\
Qwen3-4B-Base & {\cellcolor[HTML]{FDFEDA}}0.18 & {\cellcolor[HTML]{FDFEDA}}0.19 & {\cellcolor[HTML]{FDFEDA}}0.19 & {\cellcolor[HTML]{FEFFDE}}0.16 & {\cellcolor[HTML]{FDFEDB}}0.18 & {\cellcolor[HTML]{FDFEDD}}0.17 & {\cellcolor[HTML]{FFFFE5}}0.13 & {\cellcolor[HTML]{FDFEDB}}0.18 & {\cellcolor[HTML]{FEFFE1}}0.15 & {\cellcolor[HTML]{FEFFE2}}0.15 & {\cellcolor[HTML]{FAFDCC}}0.26 & {\cellcolor[HTML]{FCFED3}}0.22 & {\cellcolor[HTML]{FAFDCB}}0.27 & {\cellcolor[HTML]{FCFED3}}0.22 & {\cellcolor[HTML]{FCFED7}}0.20 & {\cellcolor[HTML]{FFFFE5}}{\cellcolor{white}}0.11 & {\cellcolor[HTML]{FFFFE5}}{\cellcolor{white}}0.10 & {\cellcolor[HTML]{FFFFE5}}{\cellcolor{white}}0.10 & {\cellcolor[HTML]{FEFFDF}}0.16 & {\cellcolor[HTML]{FCFED6}}0.21 & {\cellcolor[HTML]{F8FDC1}}0.32 & \bfseries {\cellcolor[HTML]{F1FAB5}}0.41 & {\cellcolor[HTML]{FDFED9}}0.19 \\
RoBERTa-large & {\cellcolor[HTML]{FFFFE4}}0.14 & {\cellcolor[HTML]{FEFFE1}}0.15 & {\cellcolor[HTML]{FFFFE4}}0.14 & {\cellcolor[HTML]{FFFFE5}}{\cellcolor{white}}0.12 & {\cellcolor[HTML]{FFFFE5}}0.13 & {\cellcolor[HTML]{FFFFE4}}0.13 & {\cellcolor[HTML]{FFFFE5}}0.13 & {\cellcolor[HTML]{FEFFE2}}0.15 & {\cellcolor[HTML]{FDFEDD}}0.17 & {\cellcolor[HTML]{FFFFE5}}{\cellcolor{white}}0.12 & {\cellcolor[HTML]{FDFEDD}}0.17 & {\cellcolor[HTML]{FCFED7}}0.20 & {\cellcolor[HTML]{FFFFE5}}{\cellcolor{white}}0.07 & {\cellcolor[HTML]{FFFFE5}}{\cellcolor{white}}0.09 & {\cellcolor[HTML]{FFFFE5}}{\cellcolor{white}}0.06 & {\cellcolor[HTML]{FFFFE5}}{\cellcolor{white}}0.09 & {\cellcolor[HTML]{FFFFE5}}{\cellcolor{white}}0.12 & {\cellcolor[HTML]{FFFFE5}}{\cellcolor{white}}0.10 & {\cellcolor[HTML]{FFFFE5}}{\cellcolor{white}}0.12 & {\cellcolor[HTML]{FEFFE2}}0.14 & {\cellcolor[HTML]{FFFFE5}}{\cellcolor{white}}0.12 & {\cellcolor[HTML]{FAFDC9}}0.28 & {\cellcolor[HTML]{FFFFE4}}0.14 \\
StatEnsemble & {\cellcolor[HTML]{FEFFE2}}0.14 & {\cellcolor[HTML]{FFFFE5}}0.13 & {\cellcolor[HTML]{FEFFE2}}0.14 & {\cellcolor[HTML]{FFFFE5}}0.13 & {\cellcolor[HTML]{FFFFE4}}0.13 & {\cellcolor[HTML]{FFFFE5}}0.13 & {\cellcolor[HTML]{FFFFE5}}{\cellcolor{white}}0.11 & {\cellcolor[HTML]{FFFFE5}}{\cellcolor{white}}0.12 & {\cellcolor[HTML]{FFFFE5}}{\cellcolor{white}}0.11 & {\cellcolor[HTML]{FFFFE5}}0.13 & {\cellcolor[HTML]{FFFFE5}}0.13 & {\cellcolor[HTML]{FFFFE5}}{\cellcolor{white}}0.09 & {\cellcolor[HTML]{FEFFDF}}0.16 & {\cellcolor[HTML]{FEFFE1}}0.15 & {\cellcolor[HTML]{FFFFE5}}0.13 & {\cellcolor[HTML]{FFFFE5}}{\cellcolor{white}}0.07 & {\cellcolor[HTML]{FFFFE5}}{\cellcolor{white}}0.06 & {\cellcolor[HTML]{FFFFE5}}{\cellcolor{white}}0.06 & {\cellcolor[HTML]{FFFFE5}}{\cellcolor{white}}0.12 & {\cellcolor[HTML]{FEFFDF}}0.16 & {\cellcolor[HTML]{FEFFDE}}0.17 & {\cellcolor[HTML]{FFFFE5}}0.13 & {\cellcolor[HTML]{FFFFE5}}0.13 \\
Fast-DetectGPT & {\cellcolor[HTML]{FFFFE5}}{\cellcolor{white}}0.11 & {\cellcolor[HTML]{FFFFE5}}{\cellcolor{white}}0.12 & {\cellcolor[HTML]{FFFFE5}}{\cellcolor{white}}0.10 & {\cellcolor[HTML]{FFFFE5}}{\cellcolor{white}}0.11 & {\cellcolor[HTML]{FFFFE5}}{\cellcolor{white}}0.11 & {\cellcolor[HTML]{FFFFE5}}{\cellcolor{white}}0.11 & {\cellcolor[HTML]{FFFFE5}}{\cellcolor{white}}0.09 & {\cellcolor[HTML]{FFFFE5}}{\cellcolor{white}}0.12 & {\cellcolor[HTML]{FFFFE5}}{\cellcolor{white}}0.11 & {\cellcolor[HTML]{FFFFE5}}{\cellcolor{white}}0.10 & {\cellcolor[HTML]{FFFFE5}}{\cellcolor{white}}0.08 & {\cellcolor[HTML]{FFFFE5}}{\cellcolor{white}}0.10 & {\cellcolor[HTML]{FFFFE5}}{\cellcolor{white}}0.11 & {\cellcolor[HTML]{FFFFE5}}0.13 & {\cellcolor[HTML]{FFFFE5}}{\cellcolor{white}}0.12 & {\cellcolor[HTML]{FFFFE5}}{\cellcolor{white}}0.12 & {\cellcolor[HTML]{FEFFDF}}0.16 & {\cellcolor[HTML]{FFFFE5}}{\cellcolor{white}}0.08 & {\cellcolor[HTML]{FFFFE5}}{\cellcolor{white}}0.08 & {\cellcolor[HTML]{FFFFE5}}{\cellcolor{white}}0.08 & {\cellcolor[HTML]{FFFFE5}}{\cellcolor{white}}0.09 & {\cellcolor[HTML]{FFFFE5}}{\cellcolor{white}}0.11 & {\cellcolor[HTML]{FFFFE5}}{\cellcolor{white}}0.11 \\
Binoculars & {\cellcolor[HTML]{FFFFE5}}{\cellcolor{white}}0.11 & {\cellcolor[HTML]{FFFFE5}}{\cellcolor{white}}0.10 & {\cellcolor[HTML]{FFFFE5}}{\cellcolor{white}}0.10 & {\cellcolor[HTML]{FFFFE5}}{\cellcolor{white}}0.07 & {\cellcolor[HTML]{FFFFE5}}{\cellcolor{white}}0.11 & {\cellcolor[HTML]{FFFFE5}}{\cellcolor{white}}0.09 & {\cellcolor[HTML]{FFFFE5}}{\cellcolor{white}}0.08 & {\cellcolor[HTML]{FFFFE5}}{\cellcolor{white}}0.08 & {\cellcolor[HTML]{FFFFE5}}{\cellcolor{white}}0.09 & {\cellcolor[HTML]{FFFFE5}}{\cellcolor{white}}0.10 & {\cellcolor[HTML]{FFFFE5}}{\cellcolor{white}}0.07 & {\cellcolor[HTML]{FFFFE5}}{\cellcolor{white}}0.09 & {\cellcolor[HTML]{FFFFE5}}{\cellcolor{white}}0.09 & {\cellcolor[HTML]{FFFFE5}}0.13 & {\cellcolor[HTML]{FFFFE5}}{\cellcolor{white}}0.08 & {\cellcolor[HTML]{FFFFE5}}{\cellcolor{white}}0.08 & {\cellcolor[HTML]{FFFFE5}}{\cellcolor{white}}0.06 & {\cellcolor[HTML]{FFFFE5}}{\cellcolor{white}}0.06 & {\cellcolor[HTML]{FFFFE5}}{\cellcolor{white}}0.07 & {\cellcolor[HTML]{FFFFE5}}0.13 & {\cellcolor[HTML]{FFFFE5}}0.13 & {\cellcolor[HTML]{FFFFE5}}{\cellcolor{white}}0.10 & {\cellcolor[HTML]{FFFFE5}}{\cellcolor{white}}0.10 \\
\midrule
\textit{Average} & {\cellcolor[HTML]{FDFEDB}}0.18 & {\cellcolor[HTML]{FEFFDE}}0.17 & {\cellcolor[HTML]{FEFFE1}}0.15 & {\cellcolor[HTML]{FEFFE1}}0.15 & {\cellcolor[HTML]{FEFFE1}}0.15 & {\cellcolor[HTML]{FEFFDF}}0.16 & {\cellcolor[HTML]{FFFFE4}}0.13 & {\cellcolor[HTML]{FEFFDF}}0.16 & {\cellcolor[HTML]{FEFFE2}}0.14 & {\cellcolor[HTML]{FFFFE4}}0.14 & {\cellcolor[HTML]{FDFEDD}}0.17 & {\cellcolor[HTML]{FDFEDB}}0.18 & {\cellcolor[HTML]{FDFEDB}}0.18 & {\cellcolor[HTML]{FDFEDA}}0.18 & {\cellcolor[HTML]{FEFFDE}}0.16 & {\cellcolor[HTML]{FFFFE5}}{\cellcolor{white}}0.12 & {\cellcolor[HTML]{FFFFE5}}{\cellcolor{white}}0.12 & {\cellcolor[HTML]{FFFFE5}}{\cellcolor{white}}0.12 & {\cellcolor[HTML]{FEFFDF}}0.16 & {\cellcolor[HTML]{FDFED9}}0.19 & {\cellcolor[HTML]{FCFED4}}0.22 & {\cellcolor[HTML]{FAFDCB}}0.27 & {\cellcolor[HTML]{FEFFDE}}0.16 \\
\bottomrule
Writing script $\rightarrow$ & Lat & Lat & Lat & Lat & Lat & Lat & Lat & Lat & Lat & Lat & Lat & Lat & Cyr & Cyr & Cyr & Lat & Lat & Lat & Lat & Grk & Arab & Han & \\
\end{tabular}
}
\caption{Per-language macro-averaged $F_1$ scores of the selected methods on MultiSocial test data (i.e., generalization to social-media domain). Abbreviations of writing scripts are as follows: Lat = Latin, Cyr = Cyrillic, Grk = Greek, Arab = Arabic, Han = Hanzi. Bolded values indicate the best method for each test language. Darker shades of green indicate higher scores. * denotes the new languages.} 
\label{tab:crossdomainresults}
\end{table*}

\section{Per-language-family cross-lingual performance}
\label{app:per-lang-family}
Table \ref{crosslingualresultsperfamily} provides details on the per-language-family cross-lingual transferability by aggregating the single-language cross-lingual transferability results from  Table \ref{crosslingualresults} in the main text by language family.

Table~\ref{crosslingualresultsslavic} provides cross-lingual transferability of three selected detection methods fine-tuned on individual Slavic languages, analogously to Table~\ref{crosslingualresults}. We can see a similarly strong transferability as in case of the Russian language.

\begin{table*}[!t]
\centering
\setlength{\tabcolsep}{1.0mm}
\scalebox{0.8}{
\begin{tabular}{l|l|c|c|c|c|c|c|c|c}
 \multicolumn{2}{r|}{Lang. family $\rightarrow$} & Germanic & Romance & Slavic-Latin & Slavic-Cyrillic & Uralic & Greek & Semitic & Sino-Tibetan \\
 \midrule
 & Method $\downarrow$ & ($N=3$) & ($N=3$) & ($N=5$) & ($N=3$) & ($N=1$) & ($N=1$) & ($N=1$) & ($N=1$) \\
 \toprule
\multirow[c]{6}{*}{en} & Qwen3-4B-Base & {\cellcolor[HTML]{E3F4AA}}0.52 & {\cellcolor[HTML]{F1FAB5}}0.41 & {\cellcolor[HTML]{FCFED3}}0.23 & {\cellcolor[HTML]{FBFED2}}0.23 & {\cellcolor[HTML]{FFFFE5}}{\cellcolor{white}}0.12 & {\cellcolor[HTML]{FFFFE5}}{\cellcolor{white}}0.09 & {\cellcolor[HTML]{FCFED6}}0.21 & {\cellcolor[HTML]{FFFFE5}}{\cellcolor{white}}0.09 \\
 & mdok & {\cellcolor[HTML]{D9F0A3}}0.60 & {\cellcolor[HTML]{E6F5AC}}0.49 & {\cellcolor[HTML]{FAFDCB}}0.27 & {\cellcolor[HTML]{FDFEDB}}0.18 & {\cellcolor[HTML]{FAFDCC}}0.26 & {\cellcolor[HTML]{FFFFE5}}{\cellcolor{white}}0.10 & {\cellcolor[HTML]{FFFFE5}}{\cellcolor{white}}0.10 & {\cellcolor[HTML]{FFFFE5}}{\cellcolor{white}}0.12 \\
 & OTBDetector & {\cellcolor[HTML]{DBF1A4}}0.58 & {\cellcolor[HTML]{E7F6AD}}0.49 & {\cellcolor[HTML]{F3FAB6}}0.39 & {\cellcolor[HTML]{F7FCB9}}0.36 & {\cellcolor[HTML]{F8FCBD}}0.34 & {\cellcolor[HTML]{FAFDC8}}0.29 & {\cellcolor[HTML]{FAFDC9}}0.27 & {\cellcolor[HTML]{FBFDCF}}0.25 \\
 & XLM-R-large & {\cellcolor[HTML]{E4F4AB}}0.51 & {\cellcolor[HTML]{F4FBB7}}0.39 & {\cellcolor[HTML]{F9FDC2}}0.31 & {\cellcolor[HTML]{F8FCC0}}0.32 & {\cellcolor[HTML]{FAFDCB}}0.27 & {\cellcolor[HTML]{FCFED4}}0.21 & {\cellcolor[HTML]{FCFED6}}0.21 & {\cellcolor[HTML]{FEFFDF}}0.16 \\
 & RoBERTa-large & {\cellcolor[HTML]{FAFDCB}}0.27 & {\cellcolor[HTML]{FFFFE5}}{\cellcolor{white}}0.09 & {\cellcolor[HTML]{FFFFE5}}{\cellcolor{white}}0.07 & {\cellcolor[HTML]{FFFFE5}}{\cellcolor{white}}0.04 & {\cellcolor[HTML]{FFFFE5}}{\cellcolor{white}}0.06 & {\cellcolor[HTML]{FFFFE5}}{\cellcolor{white}}0.05 & {\cellcolor[HTML]{FFFFE5}}{\cellcolor{white}}0.05 & {\cellcolor[HTML]{FFFFE5}}{\cellcolor{white}}0.05 \\
 & StatEnsemble & {\cellcolor[HTML]{F9FDC2}}0.31 & {\cellcolor[HTML]{FDFED9}}0.20 & {\cellcolor[HTML]{FFFFE5}}{\cellcolor{white}}0.09 & {\cellcolor[HTML]{FFFFE5}}{\cellcolor{white}}0.08 & {\cellcolor[HTML]{FFFFE5}}{\cellcolor{white}}0.11 & {\cellcolor[HTML]{FFFFE5}}{\cellcolor{white}}0.03 & {\cellcolor[HTML]{FEFFDF}}0.16 & {\cellcolor[HTML]{FDFED9}}0.19 \\
\midrule
\multirow[c]{6}{*}{es} & Qwen3-4B-Base & {\cellcolor[HTML]{C5E89A}}0.70 & {\cellcolor[HTML]{B1DF90}}0.81 & {\cellcolor[HTML]{E3F4AA}}0.52 & {\cellcolor[HTML]{E8F6AE}}0.48 & {\cellcolor[HTML]{EDF8B2}}0.43 & {\cellcolor[HTML]{FCFED3}}0.23 & {\cellcolor[HTML]{F8FDC1}}0.32 & {\cellcolor[HTML]{FFFFE5}}{\cellcolor{white}}0.12 \\
 & mdok & {\cellcolor[HTML]{CFEC9E}}0.65 & {\cellcolor[HTML]{B5E092}}0.79 & {\cellcolor[HTML]{E8F6AE}}0.48 & {\cellcolor[HTML]{F4FBB7}}0.38 & {\cellcolor[HTML]{EDF8B2}}0.43 & {\cellcolor[HTML]{FDFED9}}0.19 & {\cellcolor[HTML]{FCFED6}}0.21 & {\cellcolor[HTML]{FCFED7}}0.20 \\
 & OTBDetector & {\cellcolor[HTML]{D3EDA0}}0.63 & {\cellcolor[HTML]{BCE395}}0.75 & {\cellcolor[HTML]{E3F4AA}}0.51 & {\cellcolor[HTML]{E2F4AA}}0.52 & {\cellcolor[HTML]{F5FBB8}}0.38 & {\cellcolor[HTML]{F7FCBA}}0.36 & {\cellcolor[HTML]{F2FAB5}}0.40 & {\cellcolor[HTML]{F8FCC0}}0.32 \\
 & XLM-R-large & {\cellcolor[HTML]{E5F5AC}}0.50 & {\cellcolor[HTML]{E4F4AB}}0.51 & {\cellcolor[HTML]{F7FCBC}}0.35 & {\cellcolor[HTML]{F7FCBA}}0.36 & {\cellcolor[HTML]{F9FDC2}}0.31 & {\cellcolor[HTML]{FAFDCC}}0.26 & {\cellcolor[HTML]{FBFDCE}}0.25 & {\cellcolor[HTML]{FCFED7}}0.21 \\
 & RoBERTa-large & {\cellcolor[HTML]{F2FAB5}}0.40 & {\cellcolor[HTML]{D9F0A3}}0.59 & {\cellcolor[HTML]{F9FDC4}}0.30 & {\cellcolor[HTML]{FFFFE5}}{\cellcolor{white}}0.04 & {\cellcolor[HTML]{FDFED9}}0.19 & {\cellcolor[HTML]{FFFFE5}}{\cellcolor{white}}0.06 & {\cellcolor[HTML]{FFFFE5}}{\cellcolor{white}}0.07 & {\cellcolor[HTML]{FFFFE5}}{\cellcolor{white}}0.07 \\
 & StatEnsemble & {\cellcolor[HTML]{F7FCBC}}0.35 & {\cellcolor[HTML]{F1FAB5}}0.40 & {\cellcolor[HTML]{FDFEDA}}0.19 & {\cellcolor[HTML]{FBFED2}}0.23 & {\cellcolor[HTML]{FAFDC8}}0.28 & {\cellcolor[HTML]{FBFED0}}0.24 & {\cellcolor[HTML]{FBFDCE}}0.26 & {\cellcolor[HTML]{FDFED9}}0.20 \\
\midrule
\multirow[c]{6}{*}{ru} & Qwen3-4B-Base & {\cellcolor[HTML]{DDF1A6}}0.56 & {\cellcolor[HTML]{CEEB9E}}0.65 & {\cellcolor[HTML]{BAE394}}0.76 & {\cellcolor[HTML]{AFDE8F}}0.81 & {\cellcolor[HTML]{E6F5AC}}0.50 & {\cellcolor[HTML]{EDF8B2}}0.44 & \bfseries {\cellcolor[HTML]{E4F4AB}}0.51 & {\cellcolor[HTML]{FAFDCC}}0.26 \\
 & mdok & {\cellcolor[HTML]{D3EDA0}}0.62 & {\cellcolor[HTML]{C7E89A}}0.69 & \bfseries {\cellcolor[HTML]{B9E294}}0.77 & \bfseries {\cellcolor[HTML]{AEDD8E}}0.83 & \bfseries {\cellcolor[HTML]{C5E89A}}0.70 & {\cellcolor[HTML]{F4FBB7}}0.38 & {\cellcolor[HTML]{EFF9B3}}0.42 & {\cellcolor[HTML]{F8FDC1}}0.32 \\
 & OTBDetector & {\cellcolor[HTML]{E1F3A9}}0.53 & {\cellcolor[HTML]{E8F6AE}}0.48 & {\cellcolor[HTML]{BEE596}}0.74 & {\cellcolor[HTML]{B2DF90}}0.80 & {\cellcolor[HTML]{CEEB9E}}0.65 & \bfseries {\cellcolor[HTML]{E7F6AD}}0.49 & {\cellcolor[HTML]{EAF7AF}}0.46 & \bfseries {\cellcolor[HTML]{ECF7B1}}0.45 \\
 & XLM-R-large & {\cellcolor[HTML]{F8FDC1}}0.32 & {\cellcolor[HTML]{F7FCBC}}0.35 & {\cellcolor[HTML]{CFEC9E}}0.65 & {\cellcolor[HTML]{CCEA9D}}0.66 & {\cellcolor[HTML]{D0EC9F}}0.64 & {\cellcolor[HTML]{F3FAB6}}0.40 & {\cellcolor[HTML]{EEF9B3}}0.43 & {\cellcolor[HTML]{F7FCB9}}0.37 \\
 & RoBERTa-large & {\cellcolor[HTML]{FFFFE5}}{\cellcolor{white}}0.06 & {\cellcolor[HTML]{FFFFE5}}{\cellcolor{white}}0.07 & {\cellcolor[HTML]{FFFFE5}}{\cellcolor{white}}0.08 & {\cellcolor[HTML]{F2FAB5}}0.40 & {\cellcolor[HTML]{FFFFE5}}{\cellcolor{white}}0.06 & {\cellcolor[HTML]{FCFED3}}0.22 & {\cellcolor[HTML]{FEFFE1}}0.15 & {\cellcolor[HTML]{FFFFE5}}{\cellcolor{white}}0.09 \\
 & StatEnsemble & {\cellcolor[HTML]{FBFDCF}}0.24 & {\cellcolor[HTML]{FBFDCE}}0.25 & {\cellcolor[HTML]{FAFDCC}}0.26 & {\cellcolor[HTML]{EEF9B3}}0.43 & {\cellcolor[HTML]{F0F9B4}}0.42 & {\cellcolor[HTML]{FAFDC8}}0.29 & {\cellcolor[HTML]{FAFDCC}}0.26 & {\cellcolor[HTML]{FBFED0}}0.24 \\
\midrule
\multirow[c]{6}{*}{en-es-ru} & Qwen3-4B-Base & \bfseries {\cellcolor[HTML]{B1DF90}}0.81 & {\cellcolor[HTML]{ACDD8E}}0.83 & {\cellcolor[HTML]{BAE394}}0.76 & {\cellcolor[HTML]{B9E294}}0.76 & {\cellcolor[HTML]{CFEC9E}}0.65 & {\cellcolor[HTML]{EFF9B3}}0.42 & {\cellcolor[HTML]{EEF9B3}}0.43 & {\cellcolor[HTML]{F9FDC4}}0.31 \\
 & mdok & {\cellcolor[HTML]{B1DF90}}0.81 & \bfseries {\cellcolor[HTML]{ABDC8D}}0.84 & {\cellcolor[HTML]{BEE596}}0.74 & {\cellcolor[HTML]{B3E091}}0.80 & {\cellcolor[HTML]{D0EC9F}}0.64 & {\cellcolor[HTML]{F3FAB6}}0.39 & {\cellcolor[HTML]{EBF7B0}}0.46 & {\cellcolor[HTML]{F7FCBC}}0.35 \\
 & OTBDetector & {\cellcolor[HTML]{C7E89A}}0.69 & {\cellcolor[HTML]{C0E597}}0.73 & {\cellcolor[HTML]{D2EDA0}}0.64 & {\cellcolor[HTML]{C1E698}}0.72 & {\cellcolor[HTML]{E4F4AB}}0.51 & {\cellcolor[HTML]{EFF9B3}}0.42 & {\cellcolor[HTML]{EFF9B3}}0.42 & {\cellcolor[HTML]{F8FCBD}}0.34 \\
 & XLM-R-large & {\cellcolor[HTML]{E7F6AD}}0.49 & {\cellcolor[HTML]{E6F5AC}}0.49 & {\cellcolor[HTML]{D7EFA2}}0.61 & {\cellcolor[HTML]{DDF1A6}}0.57 & {\cellcolor[HTML]{D9F0A3}}0.60 & {\cellcolor[HTML]{F7FCBA}}0.36 & {\cellcolor[HTML]{F0F9B4}}0.41 & {\cellcolor[HTML]{F8FCBE}}0.34 \\
 & RoBERTa-large & {\cellcolor[HTML]{E2F4AA}}0.52 & {\cellcolor[HTML]{E1F3A9}}0.53 & {\cellcolor[HTML]{F7FCB9}}0.36 & {\cellcolor[HTML]{F4FBB7}}0.39 & {\cellcolor[HTML]{FBFED2}}0.23 & {\cellcolor[HTML]{FBFDCF}}0.25 & {\cellcolor[HTML]{FDFEDB}}0.18 & {\cellcolor[HTML]{FEFFE1}}0.15 \\
 & StatEnsemble & {\cellcolor[HTML]{F3FAB6}}0.39 & {\cellcolor[HTML]{F4FBB7}}0.39 & {\cellcolor[HTML]{FAFDC9}}0.27 & {\cellcolor[HTML]{F8FCBD}}0.34 & {\cellcolor[HTML]{F6FCB8}}0.37 & {\cellcolor[HTML]{FAFDCB}}0.27 & {\cellcolor[HTML]{FAFDC9}}0.28 & {\cellcolor[HTML]{FCFED4}}0.22 \\
 \bottomrule
\end{tabular}
}
\caption{\textbf{(RQ2)} Per-language-family cross-lingual performance (macro $F_1$) of the selected methods on test data. Rows are grouped by training language. $N$ denotes the number of test languages belonging to the language family, from which the mean value is calculated. Bolded values correspond to the best results per test-language-group.}
\label{crosslingualresultsperfamily}
\end{table*}

\begin{table*}[!t]
\centering
\setlength{\tabcolsep}{1.0mm}
\scalebox{0.73}{
\begin{tabular}{c|l||c|c|c||c|c|c||c|c|c|c|c||c|c|c||c|c|c|c||c}
\multicolumn{2}{r||}{Lang. family $\rightarrow$} & \multicolumn{3}{c||}{Germanic} & \multicolumn{3}{c||}{Romance} & \multicolumn{5}{c||}{Slavic-Latin} & \multicolumn{3}{c||}{Slavic-Cyrillic} & \multicolumn{4}{c||}{Others} & \\
\hline
& Method $\downarrow$ & de & en & nl & es & pt & ro & cs & hr & pl & sk & sl & bg & ru & uk & hu & el & ar & zh & all \\
\hline
\multirow[c]{3}{*}{ru} & Qwen3-4B-Base & {\cellcolor[HTML]{D0EC9F}}0.64 & {\cellcolor[HTML]{F1FAB5}}0.40 & {\cellcolor[HTML]{CFEC9E}}0.65 & {\cellcolor[HTML]{D7EFA2}}0.61 & {\cellcolor[HTML]{D0EC9F}}0.64 & {\cellcolor[HTML]{C3E698}}0.72 & {\cellcolor[HTML]{B8E293}}0.77 & {\cellcolor[HTML]{BEE596}}0.73 & {\cellcolor[HTML]{BEE596}}0.74 & {\cellcolor[HTML]{B3E091}}0.79 & {\cellcolor[HTML]{BAE394}}0.76 & {\cellcolor[HTML]{B2DF90}}0.80 & {\cellcolor[HTML]{A4D98A}}0.87 & {\cellcolor[HTML]{B8E293}}0.77 & {\cellcolor[HTML]{E6F5AC}}0.50 & {\cellcolor[HTML]{EDF8B2}}0.44 & {\cellcolor[HTML]{E4F4AB}}0.51 & {\cellcolor[HTML]{FAFDCC}}0.26 & {\cellcolor[HTML]{CBEA9C}}0.67 \\
 & mdok & {\cellcolor[HTML]{C0E597}}0.73 & {\cellcolor[HTML]{E6F5AC}}0.49 & {\cellcolor[HTML]{CFEC9E}}0.65 & {\cellcolor[HTML]{D2EDA0}}0.63 & {\cellcolor[HTML]{C1E698}}0.72 & {\cellcolor[HTML]{C3E698}}0.72 & {\cellcolor[HTML]{B2DF90}}0.80 & {\cellcolor[HTML]{BCE395}}0.75 & {\cellcolor[HTML]{B8E293}}0.77 & {\cellcolor[HTML]{BEE596}}0.74 & {\cellcolor[HTML]{B5E092}}0.79 & {\cellcolor[HTML]{B2DF90}}0.80 & {\cellcolor[HTML]{A2D88A}}0.88 & {\cellcolor[HTML]{B2DF90}}0.80 & {\cellcolor[HTML]{C5E89A}}0.70 & {\cellcolor[HTML]{F4FBB7}}0.38 & {\cellcolor[HTML]{EFF9B3}}0.42 & {\cellcolor[HTML]{F8FDC1}}0.32 & {\cellcolor[HTML]{C8E99B}}0.68 \\
 & XLM-R-large & {\cellcolor[HTML]{EEF9B3}}0.43 & {\cellcolor[HTML]{FBFED2}}0.23 & {\cellcolor[HTML]{F9FDC4}}0.30 & {\cellcolor[HTML]{F9FDC4}}0.30 & {\cellcolor[HTML]{F9FDC4}}0.30 & {\cellcolor[HTML]{EDF8B1}}0.44 & {\cellcolor[HTML]{C0E597}}0.73 & {\cellcolor[HTML]{DAF0A4}}0.59 & {\cellcolor[HTML]{D5EEA1}}0.62 & {\cellcolor[HTML]{D0EC9F}}0.65 & {\cellcolor[HTML]{CBEA9C}}0.67 & {\cellcolor[HTML]{CBEA9C}}0.67 & {\cellcolor[HTML]{D2EDA0}}0.63 & {\cellcolor[HTML]{C7E89A}}0.69 & {\cellcolor[HTML]{D0EC9F}}0.64 & {\cellcolor[HTML]{F3FAB6}}0.40 & {\cellcolor[HTML]{EEF9B3}}0.43 & {\cellcolor[HTML]{F7FCB9}}0.37 & {\cellcolor[HTML]{E2F4AA}}0.53 \\
\midrule
\multirow[c]{3}{*}{bg} & Qwen3-4B-Base & {\cellcolor[HTML]{E1F3A9}}0.53 & {\cellcolor[HTML]{E7F6AD}}0.49 & {\cellcolor[HTML]{DCF1A5}}0.58 & {\cellcolor[HTML]{DDF2A6}}0.56 & {\cellcolor[HTML]{DAF0A4}}0.59 & {\cellcolor[HTML]{C0E597}}0.73 & {\cellcolor[HTML]{B8E293}}0.77 & {\cellcolor[HTML]{B9E294}}0.77 & {\cellcolor[HTML]{C0E597}}0.73 & {\cellcolor[HTML]{B8E293}}0.77 & {\cellcolor[HTML]{B8E293}}0.77 & {\cellcolor[HTML]{9AD587}}0.91 & {\cellcolor[HTML]{C3E698}}0.72 & {\cellcolor[HTML]{B2DF90}}0.81 & {\cellcolor[HTML]{CBEA9C}}0.67 & {\cellcolor[HTML]{E1F3A9}}0.53 & {\cellcolor[HTML]{E0F3A8}}0.54 & {\cellcolor[HTML]{FBFED0}}0.24 & {\cellcolor[HTML]{CBEA9C}}0.67 \\
 & mdok & {\cellcolor[HTML]{D7EFA2}}0.60 & {\cellcolor[HTML]{F7FCB9}}0.37 & {\cellcolor[HTML]{DBF1A4}}0.58 & {\cellcolor[HTML]{D9F0A3}}0.60 & {\cellcolor[HTML]{D0EC9F}}0.64 & {\cellcolor[HTML]{C0E597}}0.73 & {\cellcolor[HTML]{B2DF90}}0.80 & {\cellcolor[HTML]{BAE394}}0.76 & {\cellcolor[HTML]{BCE395}}0.75 & {\cellcolor[HTML]{B5E092}}0.79 & {\cellcolor[HTML]{ACDD8E}}0.83 & {\cellcolor[HTML]{97D385}}0.93 & {\cellcolor[HTML]{B6E192}}0.78 & {\cellcolor[HTML]{B2DF90}}0.80 & {\cellcolor[HTML]{BCE395}}0.75 & {\cellcolor[HTML]{E5F5AC}}0.50 & {\cellcolor[HTML]{F9FDC2}}0.31 & {\cellcolor[HTML]{FBFDCF}}0.24 & {\cellcolor[HTML]{C9E99C}}0.68 \\
 & XLM-R-large & {\cellcolor[HTML]{F1FAB5}}0.40 & {\cellcolor[HTML]{FBFED0}}0.24 & {\cellcolor[HTML]{F4FBB7}}0.39 & {\cellcolor[HTML]{F8FCBD}}0.34 & {\cellcolor[HTML]{F7FCB9}}0.37 & {\cellcolor[HTML]{E5F5AC}}0.50 & {\cellcolor[HTML]{C8E99B}}0.69 & {\cellcolor[HTML]{D5EEA1}}0.62 & {\cellcolor[HTML]{D6EFA2}}0.61 & {\cellcolor[HTML]{C3E698}}0.71 & {\cellcolor[HTML]{BCE395}}0.75 & {\cellcolor[HTML]{BEE596}}0.74 & {\cellcolor[HTML]{E1F3A9}}0.53 & {\cellcolor[HTML]{D3EDA0}}0.63 & {\cellcolor[HTML]{C8E99B}}0.68 & {\cellcolor[HTML]{E9F6AF}}0.47 & {\cellcolor[HTML]{E4F4AB}}0.51 & {\cellcolor[HTML]{F6FCB8}}0.37 & {\cellcolor[HTML]{DEF2A7}}0.55 \\
\midrule
\multirow[c]{3}{*}{uk} & Qwen3-4B-Base & {\cellcolor[HTML]{E7F6AD}}0.49 & {\cellcolor[HTML]{F8FCC0}}0.33 & {\cellcolor[HTML]{D5EEA1}}0.62 & {\cellcolor[HTML]{E3F4AA}}0.51 & {\cellcolor[HTML]{DCF1A5}}0.57 & {\cellcolor[HTML]{CCEA9D}}0.66 & {\cellcolor[HTML]{C1E698}}0.72 & {\cellcolor[HTML]{C7E89A}}0.69 & {\cellcolor[HTML]{C4E799}}0.71 & {\cellcolor[HTML]{C1E698}}0.73 & {\cellcolor[HTML]{C8E99B}}0.69 & {\cellcolor[HTML]{BAE394}}0.76 & {\cellcolor[HTML]{C9E99C}}0.68 & {\cellcolor[HTML]{A1D889}}0.89 & {\cellcolor[HTML]{E4F4AB}}0.51 & {\cellcolor[HTML]{EEF9B3}}0.43 & {\cellcolor[HTML]{EDF8B1}}0.44 & {\cellcolor[HTML]{FEFFDE}}0.17 & {\cellcolor[HTML]{D6EFA2}}0.61 \\
 & mdok & {\cellcolor[HTML]{D0EC9F}}0.64 & {\cellcolor[HTML]{EAF7AF}}0.47 & {\cellcolor[HTML]{C7E89A}}0.69 & {\cellcolor[HTML]{CCEA9D}}0.66 & {\cellcolor[HTML]{C5E89A}}0.70 & {\cellcolor[HTML]{B8E293}}0.77 & {\cellcolor[HTML]{AEDD8E}}0.82 & {\cellcolor[HTML]{B5E092}}0.79 & {\cellcolor[HTML]{B2DF90}}0.80 & {\cellcolor[HTML]{C0E597}}0.73 & {\cellcolor[HTML]{BCE395}}0.75 & {\cellcolor[HTML]{ACDD8E}}0.83 & {\cellcolor[HTML]{AEDD8E}}0.83 & {\cellcolor[HTML]{98D486}}0.92 & {\cellcolor[HTML]{C5E89A}}0.70 & {\cellcolor[HTML]{E8F6AE}}0.48 & {\cellcolor[HTML]{F7FCB9}}0.36 & {\cellcolor[HTML]{FBFDCF}}0.24 & {\cellcolor[HTML]{C7E89A}}0.69 \\
 & XLM-R-large & {\cellcolor[HTML]{F6FCB8}}0.37 & {\cellcolor[HTML]{FBFDCF}}0.25 & {\cellcolor[HTML]{F7FCB9}}0.36 & {\cellcolor[HTML]{FAFDC8}}0.28 & {\cellcolor[HTML]{F9FDC4}}0.30 & {\cellcolor[HTML]{EEF9B3}}0.43 & {\cellcolor[HTML]{C9E99C}}0.68 & {\cellcolor[HTML]{DCF1A5}}0.57 & {\cellcolor[HTML]{DBF1A4}}0.58 & {\cellcolor[HTML]{CCEA9D}}0.67 & {\cellcolor[HTML]{C7E89A}}0.69 & {\cellcolor[HTML]{CCEA9D}}0.67 & {\cellcolor[HTML]{DFF3A8}}0.55 & {\cellcolor[HTML]{CBEA9C}}0.67 & {\cellcolor[HTML]{D2EDA0}}0.64 & {\cellcolor[HTML]{EEF9B3}}0.43 & {\cellcolor[HTML]{E8F6AE}}0.48 & {\cellcolor[HTML]{F6FCB8}}0.37 & {\cellcolor[HTML]{E2F4AA}}0.52 \\
\midrule
\multirow[c]{3}{*}{cs} & Qwen3-4B-Base & {\cellcolor[HTML]{CFEC9E}}0.65 & {\cellcolor[HTML]{F7FCB9}}0.37 & {\cellcolor[HTML]{D7EFA2}}0.61 & {\cellcolor[HTML]{E3F4AA}}0.52 & {\cellcolor[HTML]{D2EDA0}}0.63 & {\cellcolor[HTML]{B5E092}}0.79 & {\cellcolor[HTML]{98D486}}0.92 & {\cellcolor[HTML]{B2DF90}}0.80 & {\cellcolor[HTML]{B6E192}}0.78 & {\cellcolor[HTML]{9DD688}}0.90 & {\cellcolor[HTML]{AEDD8E}}0.83 & {\cellcolor[HTML]{CEEB9E}}0.66 & {\cellcolor[HTML]{DCF1A5}}0.58 & {\cellcolor[HTML]{D0EC9F}}0.64 & {\cellcolor[HTML]{CCEA9D}}0.67 & {\cellcolor[HTML]{F6FCB8}}0.37 & {\cellcolor[HTML]{E8F6AE}}0.48 & {\cellcolor[HTML]{FDFEDB}}0.18 & {\cellcolor[HTML]{CEEB9E}}0.66 \\
 & mdok & {\cellcolor[HTML]{C7E89A}}0.69 & {\cellcolor[HTML]{F8FDC1}}0.32 & {\cellcolor[HTML]{D6EFA2}}0.61 & {\cellcolor[HTML]{E9F6AF}}0.47 & {\cellcolor[HTML]{DBF1A4}}0.58 & {\cellcolor[HTML]{BDE496}}0.74 & {\cellcolor[HTML]{95D385}}0.94 & {\cellcolor[HTML]{B2DF90}}0.81 & {\cellcolor[HTML]{AEDD8E}}0.83 & {\cellcolor[HTML]{A7DB8C}}0.85 & {\cellcolor[HTML]{AEDD8E}}0.83 & {\cellcolor[HTML]{E6F5AC}}0.49 & {\cellcolor[HTML]{DAF0A4}}0.59 & {\cellcolor[HTML]{D5EEA1}}0.62 & {\cellcolor[HTML]{BEE596}}0.74 & {\cellcolor[HTML]{FBFED0}}0.24 & {\cellcolor[HTML]{FAFDCB}}0.27 & {\cellcolor[HTML]{FCFED3}}0.23 & {\cellcolor[HTML]{D2EDA0}}0.63 \\
 & XLM-R-large & {\cellcolor[HTML]{E8F6AE}}0.48 & {\cellcolor[HTML]{FBFED2}}0.23 & {\cellcolor[HTML]{F1FAB5}}0.41 & {\cellcolor[HTML]{F8FCBE}}0.33 & {\cellcolor[HTML]{F7FCBC}}0.35 & {\cellcolor[HTML]{E8F6AE}}0.48 & {\cellcolor[HTML]{BAE394}}0.76 & {\cellcolor[HTML]{DAF0A4}}0.59 & {\cellcolor[HTML]{D2EDA0}}0.63 & {\cellcolor[HTML]{B9E294}}0.77 & {\cellcolor[HTML]{C0E597}}0.73 & {\cellcolor[HTML]{D0EC9F}}0.64 & {\cellcolor[HTML]{DFF3A8}}0.55 & {\cellcolor[HTML]{D6EFA2}}0.61 & {\cellcolor[HTML]{CCEA9D}}0.66 & {\cellcolor[HTML]{F7FCBA}}0.36 & {\cellcolor[HTML]{F0F9B4}}0.42 & {\cellcolor[HTML]{F9FDC7}}0.29 & {\cellcolor[HTML]{E0F3A8}}0.54 \\
\midrule
\multirow[c]{3}{*}{hr} & Qwen3-4B-Base & {\cellcolor[HTML]{CBEA9C}}0.67 & {\cellcolor[HTML]{FBFED0}}0.24 & {\cellcolor[HTML]{CCEA9D}}0.66 & {\cellcolor[HTML]{E7F6AD}}0.49 & {\cellcolor[HTML]{DDF2A6}}0.56 & {\cellcolor[HTML]{C1E698}}0.73 & {\cellcolor[HTML]{B9E294}}0.77 & {\cellcolor[HTML]{95D385}}0.94 & {\cellcolor[HTML]{B8E293}}0.78 & {\cellcolor[HTML]{C1E698}}0.72 & {\cellcolor[HTML]{E1F3A9}}0.53 & {\cellcolor[HTML]{CBEA9C}}0.67 & {\cellcolor[HTML]{E7F6AD}}0.49 & {\cellcolor[HTML]{DFF3A8}}0.55 & {\cellcolor[HTML]{D6EFA2}}0.62 & {\cellcolor[HTML]{F8FCBD}}0.34 & {\cellcolor[HTML]{F5FBB8}}0.38 & {\cellcolor[HTML]{FEFFDF}}0.16 & {\cellcolor[HTML]{D9F0A3}}0.60 \\
 & mdok & {\cellcolor[HTML]{C4E799}}0.71 & {\cellcolor[HTML]{EFF9B3}}0.42 & {\cellcolor[HTML]{D0EC9F}}0.64 & {\cellcolor[HTML]{E0F3A8}}0.54 & {\cellcolor[HTML]{DAF0A4}}0.59 & {\cellcolor[HTML]{B5E092}}0.79 & {\cellcolor[HTML]{B9E294}}0.76 & {\cellcolor[HTML]{92D183}}0.95 & {\cellcolor[HTML]{BEE596}}0.73 & {\cellcolor[HTML]{C4E799}}0.71 & {\cellcolor[HTML]{F8FDC1}}0.32 & {\cellcolor[HTML]{D3EDA0}}0.63 & {\cellcolor[HTML]{DBF1A4}}0.58 & {\cellcolor[HTML]{E2F4AA}}0.52 & {\cellcolor[HTML]{CFEC9E}}0.65 & {\cellcolor[HTML]{F9FDC4}}0.31 & {\cellcolor[HTML]{FBFED0}}0.24 & {\cellcolor[HTML]{FDFEDA}}0.19 & {\cellcolor[HTML]{D9F0A3}}0.59 \\
 & XLM-R-large & {\cellcolor[HTML]{F3FAB6}}0.39 & {\cellcolor[HTML]{FFFFE4}}0.13 & {\cellcolor[HTML]{F5FBB8}}0.38 & {\cellcolor[HTML]{FBFED2}}0.23 & {\cellcolor[HTML]{FAFDC8}}0.28 & {\cellcolor[HTML]{E5F5AC}}0.50 & {\cellcolor[HTML]{D2EDA0}}0.63 & {\cellcolor[HTML]{BAE394}}0.76 & {\cellcolor[HTML]{DCF1A5}}0.57 & {\cellcolor[HTML]{E4F4AB}}0.51 & {\cellcolor[HTML]{FEFFDF}}0.15 & {\cellcolor[HTML]{DDF2A6}}0.56 & {\cellcolor[HTML]{EDF8B2}}0.44 & {\cellcolor[HTML]{E5F5AC}}0.50 & {\cellcolor[HTML]{DBF1A4}}0.58 & {\cellcolor[HTML]{F7FCB9}}0.36 & {\cellcolor[HTML]{F8FCBD}}0.34 & {\cellcolor[HTML]{FAFDC8}}0.28 & {\cellcolor[HTML]{ECF7B1}}0.45 \\
\midrule
\multirow[c]{3}{*}{pl} & Qwen3-4B-Base & {\cellcolor[HTML]{D0EC9F}}0.64 & {\cellcolor[HTML]{E7F6AD}}0.49 & {\cellcolor[HTML]{CBEA9C}}0.67 & {\cellcolor[HTML]{DEF2A7}}0.56 & {\cellcolor[HTML]{D3EDA0}}0.63 & {\cellcolor[HTML]{C9E99C}}0.68 & {\cellcolor[HTML]{C0E597}}0.73 & {\cellcolor[HTML]{C1E698}}0.72 & {\cellcolor[HTML]{9DD688}}0.89 & {\cellcolor[HTML]{C3E698}}0.71 & {\cellcolor[HTML]{C7E89A}}0.69 & {\cellcolor[HTML]{DDF1A6}}0.57 & {\cellcolor[HTML]{E5F5AC}}0.50 & {\cellcolor[HTML]{DDF1A6}}0.57 & {\cellcolor[HTML]{DEF2A7}}0.55 & {\cellcolor[HTML]{F4FBB7}}0.38 & {\cellcolor[HTML]{EEF9B3}}0.43 & {\cellcolor[HTML]{FDFEDA}}0.18 & {\cellcolor[HTML]{D6EFA2}}0.61 \\
 & mdok & {\cellcolor[HTML]{C4E799}}0.71 & {\cellcolor[HTML]{F3FAB6}}0.39 & {\cellcolor[HTML]{C0E597}}0.73 & {\cellcolor[HTML]{DFF3A8}}0.55 & {\cellcolor[HTML]{D2EDA0}}0.64 & {\cellcolor[HTML]{B3E091}}0.80 & {\cellcolor[HTML]{B2DF90}}0.80 & {\cellcolor[HTML]{B2DF90}}0.80 & {\cellcolor[HTML]{98D486}}0.92 & {\cellcolor[HTML]{BEE596}}0.74 & {\cellcolor[HTML]{B3E091}}0.80 & {\cellcolor[HTML]{D7EFA2}}0.61 & {\cellcolor[HTML]{D6EFA2}}0.61 & {\cellcolor[HTML]{D0EC9F}}0.64 & {\cellcolor[HTML]{C4E799}}0.71 & {\cellcolor[HTML]{F7FCBC}}0.35 & {\cellcolor[HTML]{FAFDC9}}0.27 & {\cellcolor[HTML]{FCFED6}}0.21 & {\cellcolor[HTML]{CFEC9E}}0.65 \\
 & XLM-R-large & {\cellcolor[HTML]{E3F4AA}}0.51 & {\cellcolor[HTML]{FAFDCC}}0.26 & {\cellcolor[HTML]{E9F6AF}}0.48 & {\cellcolor[HTML]{F3FAB6}}0.39 & {\cellcolor[HTML]{EFF9B3}}0.42 & {\cellcolor[HTML]{DCF1A5}}0.57 & {\cellcolor[HTML]{C9E99C}}0.68 & {\cellcolor[HTML]{D3EDA0}}0.63 & {\cellcolor[HTML]{B8E293}}0.77 & {\cellcolor[HTML]{C1E698}}0.72 & {\cellcolor[HTML]{C8E99B}}0.68 & {\cellcolor[HTML]{D0EC9F}}0.64 & {\cellcolor[HTML]{E4F4AB}}0.51 & {\cellcolor[HTML]{DBF1A4}}0.59 & {\cellcolor[HTML]{C9E99C}}0.68 & {\cellcolor[HTML]{EDF8B2}}0.44 & {\cellcolor[HTML]{ECF7B1}}0.45 & {\cellcolor[HTML]{F1FAB5}}0.41 & {\cellcolor[HTML]{DDF2A6}}0.56 \\
\midrule
\multirow[c]{3}{*}{sk} & Qwen3-4B-Base & {\cellcolor[HTML]{E6F5AC}}0.49 & {\cellcolor[HTML]{F0F9B4}}0.41 & {\cellcolor[HTML]{E0F3A8}}0.54 & {\cellcolor[HTML]{E8F6AE}}0.48 & {\cellcolor[HTML]{D6EFA2}}0.61 & {\cellcolor[HTML]{C7E89A}}0.69 & {\cellcolor[HTML]{BAE394}}0.76 & {\cellcolor[HTML]{CEEB9E}}0.65 & {\cellcolor[HTML]{C3E698}}0.71 & {\cellcolor[HTML]{8DCF81}}0.97 & {\cellcolor[HTML]{CBEA9C}}0.67 & {\cellcolor[HTML]{D6EFA2}}0.61 & {\cellcolor[HTML]{DDF1A6}}0.57 & {\cellcolor[HTML]{DCF1A5}}0.57 & {\cellcolor[HTML]{DDF1A6}}0.56 & {\cellcolor[HTML]{EBF7B0}}0.46 & {\cellcolor[HTML]{DBF1A4}}0.58 & {\cellcolor[HTML]{FAFDC9}}0.28 & {\cellcolor[HTML]{D6EFA2}}0.61 \\
 & mdok & {\cellcolor[HTML]{DDF1A6}}0.57 & {\cellcolor[HTML]{F9FDC2}}0.31 & {\cellcolor[HTML]{E0F3A8}}0.54 & {\cellcolor[HTML]{EFF9B3}}0.42 & {\cellcolor[HTML]{E2F4AA}}0.53 & {\cellcolor[HTML]{D0EC9F}}0.64 & {\cellcolor[HTML]{BAE394}}0.76 & {\cellcolor[HTML]{C3E698}}0.71 & {\cellcolor[HTML]{C1E698}}0.72 & {\cellcolor[HTML]{8ED082}}0.96 & {\cellcolor[HTML]{BDE496}}0.74 & {\cellcolor[HTML]{DDF1A6}}0.57 & {\cellcolor[HTML]{E3F4AA}}0.51 & {\cellcolor[HTML]{E0F3A8}}0.54 & {\cellcolor[HTML]{CCEA9D}}0.67 & {\cellcolor[HTML]{F7FCBC}}0.35 & {\cellcolor[HTML]{F8FDC1}}0.32 & {\cellcolor[HTML]{FCFED3}}0.23 & {\cellcolor[HTML]{DAF0A4}}0.59 \\
 & XLM-R-large & {\cellcolor[HTML]{F1FAB5}}0.40 & {\cellcolor[HTML]{FDFEDA}}0.19 & {\cellcolor[HTML]{FAFDC8}}0.29 & {\cellcolor[HTML]{FAFDCB}}0.27 & {\cellcolor[HTML]{F4FBB7}}0.39 & {\cellcolor[HTML]{F8FCBE}}0.33 & {\cellcolor[HTML]{E3F4AA}}0.52 & {\cellcolor[HTML]{F2FAB5}}0.40 & {\cellcolor[HTML]{EBF7B0}}0.46 & {\cellcolor[HTML]{9FD788}}0.89 & {\cellcolor[HTML]{ECF7B1}}0.45 & {\cellcolor[HTML]{E7F6AD}}0.49 & {\cellcolor[HTML]{E7F6AD}}0.49 & {\cellcolor[HTML]{EEF9B3}}0.43 & {\cellcolor[HTML]{F0F9B4}}0.41 & {\cellcolor[HTML]{E8F6AE}}0.48 & {\cellcolor[HTML]{E8F6AE}}0.48 & {\cellcolor[HTML]{FBFED0}}0.24 & {\cellcolor[HTML]{ECF7B1}}0.45 \\
\midrule
\multirow[c]{3}{*}{sl} & Qwen3-4B-Base & {\cellcolor[HTML]{E0F3A8}}0.54 & {\cellcolor[HTML]{F7FCBA}}0.35 & {\cellcolor[HTML]{D9F0A3}}0.60 & {\cellcolor[HTML]{E2F4AA}}0.52 & {\cellcolor[HTML]{D0EC9F}}0.64 & {\cellcolor[HTML]{C5E89A}}0.70 & {\cellcolor[HTML]{ACDD8E}}0.83 & {\cellcolor[HTML]{B2DF90}}0.80 & {\cellcolor[HTML]{C1E698}}0.72 & {\cellcolor[HTML]{AEDD8E}}0.82 & {\cellcolor[HTML]{92D183}}0.95 & {\cellcolor[HTML]{C3E698}}0.71 & {\cellcolor[HTML]{E0F3A8}}0.54 & {\cellcolor[HTML]{D9F0A3}}0.60 & {\cellcolor[HTML]{C4E799}}0.70 & {\cellcolor[HTML]{F5FBB8}}0.38 & {\cellcolor[HTML]{F7FCBA}}0.36 & {\cellcolor[HTML]{FEFFDE}}0.16 & {\cellcolor[HTML]{D3EDA0}}0.63 \\
 & mdok & {\cellcolor[HTML]{C1E698}}0.72 & {\cellcolor[HTML]{EEF9B3}}0.43 & {\cellcolor[HTML]{C9E99C}}0.68 & {\cellcolor[HTML]{E4F4AB}}0.51 & {\cellcolor[HTML]{DFF3A8}}0.55 & {\cellcolor[HTML]{C9E99C}}0.68 & {\cellcolor[HTML]{B9E294}}0.76 & {\cellcolor[HTML]{BAE394}}0.76 & {\cellcolor[HTML]{C7E89A}}0.69 & {\cellcolor[HTML]{C1E698}}0.72 & {\cellcolor[HTML]{90D083}}0.95 & {\cellcolor[HTML]{DDF1A6}}0.57 & {\cellcolor[HTML]{E4F4AB}}0.51 & {\cellcolor[HTML]{E1F3A9}}0.53 & {\cellcolor[HTML]{D5EEA1}}0.62 & {\cellcolor[HTML]{FAFDCB}}0.27 & {\cellcolor[HTML]{FCFED6}}0.21 & {\cellcolor[HTML]{FDFEDD}}0.18 & {\cellcolor[HTML]{D9F0A3}}0.60 \\
 & XLM-R-large & {\cellcolor[HTML]{F4FBB7}}0.39 & {\cellcolor[HTML]{FBFDCE}}0.25 & {\cellcolor[HTML]{F6FCB8}}0.37 & {\cellcolor[HTML]{F8FCBE}}0.33 & {\cellcolor[HTML]{F8FCBD}}0.34 & {\cellcolor[HTML]{E7F6AD}}0.49 & {\cellcolor[HTML]{D5EEA1}}0.62 & {\cellcolor[HTML]{CBEA9C}}0.67 & {\cellcolor[HTML]{DFF3A8}}0.55 & {\cellcolor[HTML]{C5E89A}}0.70 & {\cellcolor[HTML]{BAE394}}0.76 & {\cellcolor[HTML]{D2EDA0}}0.63 & {\cellcolor[HTML]{EDF8B1}}0.45 & {\cellcolor[HTML]{E6F5AC}}0.50 & {\cellcolor[HTML]{D6EFA2}}0.61 & {\cellcolor[HTML]{F3FAB6}}0.39 & {\cellcolor[HTML]{F3FAB6}}0.39 & {\cellcolor[HTML]{F9FDC7}}0.29 & {\cellcolor[HTML]{E4F4AB}}0.51 \\
\hline
\multicolumn{2}{r||}{Writing script $\rightarrow$} & Lat & Lat & Lat & Lat & Lat & Lat & Lat & Lat & Lat & Lat & Lat & Cyr & Cyr & Cyr & Lat & Grk & Arab & Han & \\
\end{tabular}
}
\caption{\textbf{(RQ2)} Per-language cross-lingual   macro-averaged  $F_1$ scores of the selected methods fine-tuned on Slavic languages on test data. Writing scripts are as follows: Lat = Latin, Cyr = Cyrillic, Grk = Greek, Arab = Arabic, Han = Hanzi. Bolded values indicate the best method for each training-language and test-language pair. Darker shades of green indicate higher scores.
}
\label{crosslingualresultsslavic}
\end{table*}

\section{Per-generator multi-lingual and cross-lingual performance} 
\label{app:per-generator}
The finer-granularity multilingual (for each test language) results per-class (i.e., generator) of the selected authorship attribution methods are provided in Table~\ref{multilingualresults_generator}. In this single-class evaluation scenario, the performance is reported in the form of a weighted average $F_1$ score (since non-evaluated classes have no supporting samples). 
Analogously, Table~\ref{crosslingualresults_generator} reports per-generator performance of two of the best (mdok and OTBDetector) authorship attribution methods for cross-lingual experiments.

\begin{table*}[!t]
\centering
\setlength{\tabcolsep}{1.0mm}
\scalebox{0.69}{
\begin{tabular}{l|l||c|c|c||c|c|c||c|c|c|c|c||c|c|c||c|c|c|c||c}
\multicolumn{2}{r||}{Lang. family $\rightarrow$} & \multicolumn{3}{c||}{Germanic} & \multicolumn{3}{c||}{Romance} & \multicolumn{5}{c||}{Slavic-Latin} & \multicolumn{3}{c||}{Slavic-Cyrillic} & \multicolumn{4}{c||}{Others} & \\
\midrule
 & Generator (class) & de & en & nl & es & pt & ro & cs & hr & pl & sk & sl & bg & ru & uk & hu & el & ar & zh & all \\
\midrule
\multirow[c]{8}{*}{\rotatebox{90}{Qwen3-4B-Base}}& Llama-2-70b-chat-hf & {\cellcolor[HTML]{90D083}}0.96 & {\cellcolor[HTML]{8ED082}}0.96 & {\cellcolor[HTML]{8ED082}}0.96 & {\cellcolor[HTML]{93D284}}0.94 & {\cellcolor[HTML]{90D083}}0.96 & {\cellcolor[HTML]{90D083}}0.96 & {\cellcolor[HTML]{8DCF81}}0.97 & {\cellcolor[HTML]{8DCF81}}0.97 & {\cellcolor[HTML]{8BCE81}}0.98 & {\cellcolor[HTML]{8DCF81}}0.97 & {\cellcolor[HTML]{8BCE81}}0.98 & {\cellcolor[HTML]{8ED082}}0.96 & {\cellcolor[HTML]{8ED082}}0.96 & {\cellcolor[HTML]{90D083}}0.95 & {\cellcolor[HTML]{8ED082}}0.97 & {\cellcolor[HTML]{8BCE81}}0.98 & {\cellcolor[HTML]{8BCE81}}0.98 & {\cellcolor[HTML]{A4D98A}}0.87 & {\cellcolor[HTML]{8ED082}}0.96 \\
 & Mistral-7B-Instruct-v0.2 & {\cellcolor[HTML]{8ED082}}0.96 & {\cellcolor[HTML]{92D183}}0.95 & {\cellcolor[HTML]{8BCE81}}0.98 & {\cellcolor[HTML]{8DCF81}}0.97 & {\cellcolor[HTML]{8DCF81}}0.97 & {\cellcolor[HTML]{8BCE81}}0.98 & {\cellcolor[HTML]{8ED082}}0.97 & {\cellcolor[HTML]{8BCE81}}0.98 & {\cellcolor[HTML]{8BCE81}}0.98 & {\cellcolor[HTML]{8DCF81}}0.97 & {\cellcolor[HTML]{8ED082}}0.96 & {\cellcolor[HTML]{8ED082}}0.97 & {\cellcolor[HTML]{92D183}}0.95 & {\cellcolor[HTML]{98D486}}0.92 & {\cellcolor[HTML]{92D183}}0.95 & {\cellcolor[HTML]{90D083}}0.95 & {\cellcolor[HTML]{8BCE81}}0.98 & {\cellcolor[HTML]{ACDD8E}}0.83 & {\cellcolor[HTML]{90D083}}0.96 \\
\end{tabular}
}
\caption{Per-generator multi-lingual performance (weighted $F_1$) of the selected methods using the \multitude{} data.
}
\label{multilingualresults_generator}
\end{table*}

\begin{table*}[!t]
\centering
\setlength{\tabcolsep}{1.0mm}
\scalebox{0.69}{
\begin{tabular}{l|l|l||c|c|c||c|c|c||c|c|c|c|c||c|c|c||c|c|c|c||c}
& & Lang. family $\rightarrow$ & \multicolumn{3}{c||}{Germanic} & \multicolumn{3}{c||}{Romance} & \multicolumn{5}{c||}{Slavic-Latin} & \multicolumn{3}{c||}{Slavic-Cyrillic} & \multicolumn{4}{c||}{Others} & \\
\cmidrule{3-22}
& & Generator (class) & de & en & nl & es & pt & ro & cs & hr & pl & sk & sl & bg & ru & uk & hu & el & ar & zh & all \\
\midrule
\multirow[c]{16}{*}{\rotatebox{90}{en}}& \multirow[c]{8}{*}{\rotatebox{90}{mdok}}& Llama-2-70b-chat-hf & {\cellcolor[HTML]{F8FCC0}}0.32 & {\cellcolor[HTML]{90D083}}0.96 & {\cellcolor[HTML]{DAF0A4}}0.59 & {\cellcolor[HTML]{DFF3A8}}0.54 & {\cellcolor[HTML]{FFFFE5}}{\cellcolor{white}}0.12 & {\cellcolor[HTML]{D9F0A3}}0.59 & {\cellcolor[HTML]{FFFFE5}}{\cellcolor{white}}0.05 & {\cellcolor[HTML]{FBFED2}}0.23 & {\cellcolor[HTML]{F4FBB7}}0.38 & {\cellcolor[HTML]{FAFDCB}}0.27 & {\cellcolor[HTML]{FFFFE5}}{\cellcolor{white}}0.12 & {\cellcolor[HTML]{FFFFE5}}{\cellcolor{white}}0.07 & {\cellcolor[HTML]{FFFFE5}}{\cellcolor{white}}0.07 & {\cellcolor[HTML]{FDFED9}}0.19 & {\cellcolor[HTML]{FFFFE5}}{\cellcolor{white}}0.09 & {\cellcolor[HTML]{FFFFE5}}{\cellcolor{white}}0.05 & {\cellcolor[HTML]{FFFFE5}}{\cellcolor{white}}0.05 & {\cellcolor[HTML]{FFFFE5}}{\cellcolor{white}}0.08 & {\cellcolor[HTML]{F9FDC2}}0.31 \\
\end{tabular}
}
\caption{Per-generator cross-lingual performance (weighted $F_1$) of two of the best AA methods using the \multitude{} data.
}
\label{crosslingualresults_generator}
\end{table*}
               
\end{document}